\documentclass{article}
\usepackage{natbib}
\usepackage{authblk}
\usepackage{amssymb}
\usepackage{amsmath}
\usepackage{url}
\usepackage{hyperref}

\usepackage{tcolorbox}

\usepackage{footnote}
\makesavenoteenv{tabular}
\makesavenoteenv{table}

\usepackage[utf8]{inputenc}
\usepackage{textcomp} % for \textcheckmark
\usepackage{amssymb} % for \ding
\usepackage{pifont} % for \ding
\DeclareUnicodeCharacter{2713}{\checkmark}
\DeclareUnicodeCharacter{2718}{\ding{55}}

\usepackage{array} % for column width adjustments
\usepackage{adjustbox} % for resizing tables
\usepackage{graphicx} % for \resizebox

\usepackage{placeins}
\newcolumntype{L}[1]{>{\raggedright\let\newline\\\arraybackslash\hspace{0pt}}m{#1}}
\newcolumntype{C}[1]{>{\centering\let\newline\\\arraybackslash\hspace{0pt}}m{#1}}
\newcolumntype{R}[1]{>{\raggedleft\let\newline\\\arraybackslash\hspace{0pt}}m{#1}}

\title{\textbf{Trusted Knowledge Extraction \\ for Operations and Maintenance Intelligence}} %% Article title

\author{\textbf{Kathleen P. Mealey\textsuperscript{1}, Jonathan A. Karr Jr.\textsuperscript{1}, Priscila Saboia Moreira\textsuperscript{1}, Paul R. Brenner\textsuperscript{1}, 
Charles F. Vardeman II\textsuperscript{1}}} %% Author name

\affil[1]{University of Notre Dame, Notre Dame, Indiana, USA}

\affil[ ]{\texttt{kpmealey@outlook.com}, \texttt{\{jkarr, pmoreira, paul.r.brenner, cvardema\}@nd.edu}}

\date{}

\begin{document}
\maketitle
%% Abstract
\begin{abstract}
%% Text of abstract
Deriving operational intelligence from organizational data repositories is a key challenge due to the dichotomy of data confidentiality vs data integration objectives, as well as the limitations of Natural Language Processing (NLP) tools relative to the specific knowledge structure of domains such as operations and maintenance. In this work, we discuss Knowledge Graph construction and break down the Knowledge Extraction process into its Named Entity Recognition, Coreference Resolution, Named Entity Linking, and Relation Extraction functional components. We then evaluate sixteen NLP tools in concert with or in comparison to the rapidly advancing capabilities of Large Language Models (LLMs). We focus on the operational and maintenance intelligence use case for trusted applications in the aircraft industry. A baseline dataset is derived from a rich public domain US Federal Aviation Administration dataset focused on equipment failures or maintenance requirements. We assess the zero-shot performance of NLP and LLM tools that can be operated within a controlled, confidential environment (no data is sent to third parties). Based on our observation of significant performance limitations, we discuss the challenges related to trusted NLP and LLM tools as well as their Technical Readiness Level for wider use in mission-critical industries such as aviation. We conclude with recommendations to enhance trust and provide our open-source curated dataset to support further baseline testing and evaluation.
\end{abstract}

%% Keywords

% Footnote without number
\renewcommand{\thefootnote}{}
\footnotetext{Link to our dataset: \url{https://zenodo.org/records/13333825}}
\renewcommand{\thefootnote}{\arabic{footnote}}

\newpage
\tableofcontents
\newpage

\section{Introduction}

% 1) Providing clear background.
Organizations in domains such as aviation, manufacturing, and defense generate vast amounts of unstructured data in the form of reports, operational logs, and incident records. These databases hold key insights that can be leveraged to enhance safety procedures, predict maintenance timelines, streamline operations, and more. However, accessing and modeling such insights is challenging. Trends in these operational records, which we dub “operations and maintenance intelligence,” are fragmented among thousands of disconnected reports. The reports are often inconsistently structured, and meaningful knowledge is often obscured by industry shorthand and lack of context.

% 2) Defining the research problem.
One process with great potential to harness insights in large databases is Knowledge Extraction (KE), in which targeted data points are extracted and captured in a structured form, such as a Knowledge Graph (KG). In a KG, individual data entities are represented as nodes, with edges capturing the semantic relationships between them. Structured data is key for providing operations and maintenance intelligence because it is much more readily searched, analyzed, and verified than unstructured text. There are many effective open-source KE tools available out-of-the-box; however, they are trained on open-domain, conventional prose, and therefore struggle to adapt to the strange vocabulary and syntax used in operations and maintenance records. Organizations require KE tools that are both effective and trustworthy in this context, where trust includes the ability to process their data collections at acceptable levels of accuracy, understandability, robustness, reproducibility, and confidentiality.

\begin{figure}[ht!]
    \centering
    \includegraphics[width=\textwidth]{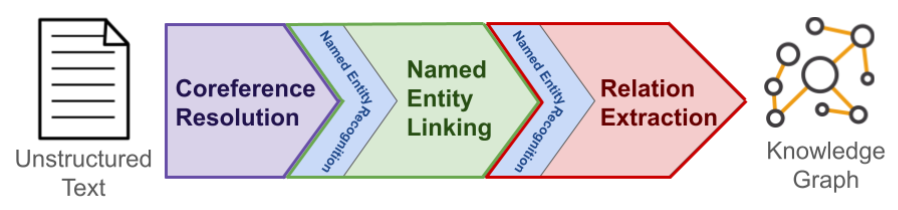}
    \caption{KE Workflow.
    The Knowledge Extraction Workflow is an approach to extracting graphical data from unstructured text. It begins with CR, which identifies different words or phrases that refer to the same entity. Then, in NEL, entities are recognized (NER) and linked to corresponding unique IDs in an external KB. Lastly, in RE, entities are recognized (NER) and connected through well-defined relationships.}
    \label{fig:KE_Workflow}
\end{figure}

% 3) Summarizing existing work approaches to address the problem.
Existing research in trustworthy KE has produced robust capabilities in several Natural Language Processing (NLP) techniques. NLP-based methods include Named Entity Recognition (NER), which identifies named entities in text and classifies them into a set of entity types \cite{sundheim_overview_1995}. In this study, we adopt a multi-stage KE workflow, based on the Information Extraction Pipeline (\cite{bratanic2021extraction}), which consists of four core NLP tasks: Coreference Resolution (CR), which links different expressions referring to the same entity and has been shown to increase accuracy in many NLP tasks (\cite{sukthanker2020anaphora}); Named Entity Linking (NEL), which enriches identified entities by linking them to unique identities in external knowledge bases\footnote{NER is often conflated with Named Entity Disambiguation (NED), which is the task of disambiguating an entity from its possible references, without necessarily linking it to an external KB (\cite{al-moslmi_named_2020}).}; and Relation Extraction (RE), which identifies meaningful relationships between entities. NER is the fourth task, which is often embedded in NEL and RE, since many NEL and RE systems utilize a multi-stage approach using an NER sub-module to extract entities before linking or relating them, respectively. Together, these steps support the construction of knowledge graphs.
The diagram in Figure~\ref{fig:KE_Workflow} summarizes the KE Workflow in our paper.

These and other NLP-based methods have been proven on several open-domain benchmark datasets. However, applying these KE capabilities to the operations and maintenance domain remains underexplored in open-source literature. Much of the work in this area is constrained by proprietary datasets, limiting reproducibility and public evaluation. As a result, research efforts are often siloed across organizations with differing data standards, workflows, and infrastructure—making it difficult to compare methods or build on shared benchmarks. A few initiatives, such as MaintNet \cite{akhbardeh2020maintnet}, have begun to address this gap by releasing annotated datasets in the aviation and automotive maintenance domains. However, such resources remain limited in scope, and there is still a need for comprehensive benchmarks to evaluate tool performance in real-world maintenance settings.

% 4) Explaining why the problem remains important.
As organizations increasingly rely on data-driven approaches for decision support, the inability to extract accurate, structured knowledge from unstructured records buries potentially critical insights. In the maintenance and operations domain, these insights could build systems for safety assurance, performance monitoring, and predictive maintenance, and more. Without trusted KE tools tailored to their specific domain, these organizations face greater operational risk, increased manual overhead, and missed opportunities for insight-driven optimization.

% 5) Describing the proposed solution.
To address this gap, we introduce the Operations and Maintenance Intelligence benchmark, or OMIn, a novel benchmark dataset in the operations and maintenance domain. OMIn is based on curated records from the publicly available FAA Accident/Incident datasets, which shares several peculiarities found in maintenance data: prevalence of rare entities, uncommon or incorrect syntax due to shorthand, abbreviations, acronyms, and small record size. We also release gold standard annotations for NER, CR, and NEL based on OMIn. Note that we did not create a gold standard for RE, for reasons discussed in Section \ref{sec:RE_GS}.

We then used OMIn to benchmark sixteen openly available NLP tools on the operations and maintenance domain in a zero-shot setting. While the term zero-shot is used in multiple ways in the field of machine learning, for the purpose of our paper, evaluation in a zero-shot setting means that none of the tools chosen have been fine-tuned on FAA data nor have been trained for the aviation or maintenance domains. We present our results to inform the selection of off-the-shelf models and identify candidates for domain adaptation or fine-tuning to meet operational requirements.

%The diagram in Figure~\ref{fig:KE_Workflow_Example} demonstrates the KE workflow performed on a sample from OMIn.

%\begin{figure}[ht!]
%    \centering
%    \includegraphics[width=\textwidth]{figures/Figure_2.png}
%    \caption{KE Workflow Applied Example.
%    Here, the KE Workflow is applied to the sentence on the bottom from OMIn to generate the graph on the top. Named entities, like \textit{aircraft} and \textit{Pilot}, are denoted in blue to signify the NER subtask in NEL and RE. Then, the CR system (purple) recognizes that the \textit{aircraft} refers to the same entity in different parts of the sentence, ensuring information relating to the aircraft is consolidated around one node. NEL (green) connects recognized entities to their corresponding Wikidata entries, such as \textit{aircraft} (Q11436) and \textit{aircraft pilot} (Q2095549). Finally, RE (red) identifies relationships between entities, with the red edges representing Wikidata properties, such as the pilot \textit{operating} the aircraft or the brake being a \textit{part of} the aircraft.}
    \label{fig:KE_Workflow_Example}
%\end{figure}

% 6) Stating the main contributions.

\paragraph{The key contributions of this work are threefold:}
\begin{itemize}
\item{} A publicly released benchmark dataset (OMIn) with gold standards for KE in the maintenance domain.
\item{} A comprehensive zero-shot evaluation of sixteen KE tools using OMIn.
\item{} An analysis of tool performance, limitations, and implications for trusted decision support in maintenance operations.
\end{itemize}

While we focus on maintenance, this technology has use-cases in many fields, such as healthcare, law, and logistics, where it is important to sort through massive amounts of unstructured data and identify patterns of interest.

% 7) A closing paragraph outlining the structure of the paper.
The remainder of this paper is structured as follows: Section 2 reviews related work. Section 3 captures the research problem. Section 4 presents the methodology used in this study, starting with the creation of the OMIn dataset, including gold standards for annotation, then KE tools selection and zero-shot implementation, and finally our evaluation methodology. Section 5 presents the results from evaluating the selected tools on the OMIn dataset. Section 6 discusses the performance, trustworthiness, and technological readiness of the surveyed tools for the maintenance domain based on the results in Section 5. We conclude by summarizing key contributions and listing suggestions for future work.

\section{Related Work}

The integration of LLMs and KGs is becoming a key area of innovation (\cite{khorashadizadeh2024research}). Together, they create potential for powerful synergy because, while LLMs excel at tasks which require a human-like ability to reason and respond, they often struggle to recall specific facts from their training data. KGs, in turn, may be used to capture such facts -- representing knowledge for LLMs to retrieve and for humans to verify and update. KE, a crucial step in creating and updating KGs, is particularly relevant for specialized domains where precise information recall and reasoning are paramount. Given the often fragmented and domain-specific nature of data in many critical industries, effective KE is essential. This integration and its applications lead us to three main points: (1) LLM-KG Advancements for Question-Answering, (2) KE in Maintenance and Aviation Domains, and (3) Open Data and Collaboration in Technical Domains.

\subsection{LLM-KG Advancements for Question-Answering}
The field of LLM-KG integration and question-answering has seen several advancements. Graph RAG offers an approach to question-answering by populating a KG with facts from text for consistent querying. This method has been shown to outperform Naïve RAG in terms of comprehensiveness, diversity, and empowerment, all while using fewer tokens, ultimately providing better contextual understanding and query scalability (\cite{edge2024localglobalgraphrag}). Furthermore, Ontology-Based Information Extraction (OBIE) has emerged as a subfield that integrates ontologies into the Information Extraction (IE) process to formally specify which concepts to extract, with existing surveys providing a taxonomy of state-of-the-art OBIE systems (\cite{KONYS20182208}). Recent work also explores how LLMs can be directly fine-tuned to extract structured information from text and populate KGs, often leveraging prompt engineering and few-shot learning to guide the extraction process for specific schema types (\cite{liu2025large}). This enables more flexible and adaptable KG construction from diverse text sources.

It is also important that these techniques work with sensitive data. The Llamdex framework (Large LAnguage Model with Domain EXpert) utilizes a private model for domain knowledge (\cite{wu2024model}). Through their secure transfer generation, they create accurate responses to domain-specific questions while preserving KE. During this process, it is important to focus on secure LLM deployment strategies, including: on-premises LLMs, secure RAG, sandboxing, data anonymization, PII scrubbing, differential privacy, access control, encryption, logging, and red-teaming strategies (\cite{matviishyn2025llm}). Privacy is important since there can be challenges with LLM training data privacy, insecure user prompts, vulnerabilities with LLM-generated outputs, and issues with LLM agents (\cite{shanmugarasa2025sok}).

\subsection{KE in Maintenance and Aviation Domains}
KE in the maintenance and aviation domains is critical, as large amounts of unstructured textual data often contain vital insights. Various methods have been explored for the construction of KGs reguarding the representation of domain knowledge (\cite{mishra_domain-targeted_2017, yu2020}). Approaches to maintenance KE include rules-based methods for extracting entities and relations, which can then be augmented by LLMs for additional contextual extraction (\cite{Dixit2021ExtractingSF}). The issue of obtaining maintenance information has been framed as a key to finding the root cause of failures, with proposed methods for classifying maintenance records based on latent semantic analysis and SVM (\cite{sharp2017}). User-friendly tools such as KNOWO have been developed to help create KGs from maintenance work orders, featuring the automatic extraction of concepts belonging to controlled vocabularies (\cite{ameri2022}). Additionally, KGs have been constructed using datasets of aircraft maintenance information, focusing on components, fault information, and maintenance measures (\cite{yue2022}). The complexity of aviation maintenance records often contain technical jargon, abbreviations, and informal language, presents unique challenges that domain-specific NLP models and specialized ontologies are increasingly addressing to improve extraction accuracy (\cite{liu2025llm}).

\subsection{Open Data and Collaboration in Technical Domains}
Maintainers and analysts have been proposed to collaborate towards developing standards for textual analysis, including entity typing, which could form the basis for technical language processing (TLP) \cite{brundage2021technical}. Initiatives like MaintNet provide collections of annotated logbook datasets across aviation, automotive, and facility maintenance domains, with further work expanding on these efforts \cite{akhbardeh2020maintnet, akhbardeh-etal-2020-nlp}. The transformation of text into a KG can be achieved through Information Extraction pipelines, which typically consist of stages such as CR, NEL, RE, and KG construction (\cite{bratanic2021extraction}). Beyond datasets, the development of shared ontologies and standardized data formats between different organizations and research groups is crucial to fostering interoperability and accelerating research in technical domains (\cite{meng2023fault}). However, it remains difficult for models to paraphrase information when it comes from domains with few resources. (\cite{li2024learning}). Therefore, it is import to have tools such as EvalxNLP which highlight how much information can be explained (\cite{dhaini2025evalxnlp}).

\section{Research Problem}

Organizations in the aviation and manufacturing domains generate vast amounts of unstructured maintenance records rich with operational insights, yet these records are difficult to analyze due to their inconsistent structure, jargon, and shorthand. KE techniques, such as CR, NER, NEL, and RE, can transform this text into structured formats like knowledge graphs, but most tools are trained on open-domain prose and perform poorly on technical, domain-specific text. Furthermore, prior research highlights the strengths and limitations of current approaches to KE, particularly in the integration of LLMs with KGs. Although LLMs excel in generalizability, they often falter in domain-specific accuracy. This gap presents a major barrier to deploying trusted KE systems in safety-critical settings, where accuracy, reproducibility, and transparency are essential. Progress is further hindered by the lack of public, domain-specific benchmarks; most datasets are proprietary or too narrow for robust evaluation. This work addresses that gap in the aviation and maintenance sectors by introducing the Operations and Maintenance Intelligence benchmark, or OMIn, a benchmark dataset based on FAA incident reports, enabling targeted evaluation of KE tools. In line with open data initiatives (\cite{brundage2021technical}), we contribute an open-source dataset and detailed evaluations for zero-shot performance of sixteen KE tools, setting a benchmark for future research and tool development in this domain.

\section{Methodology}

\begin{figure}[ht!]
    \centering
    \includegraphics[width=\textwidth]{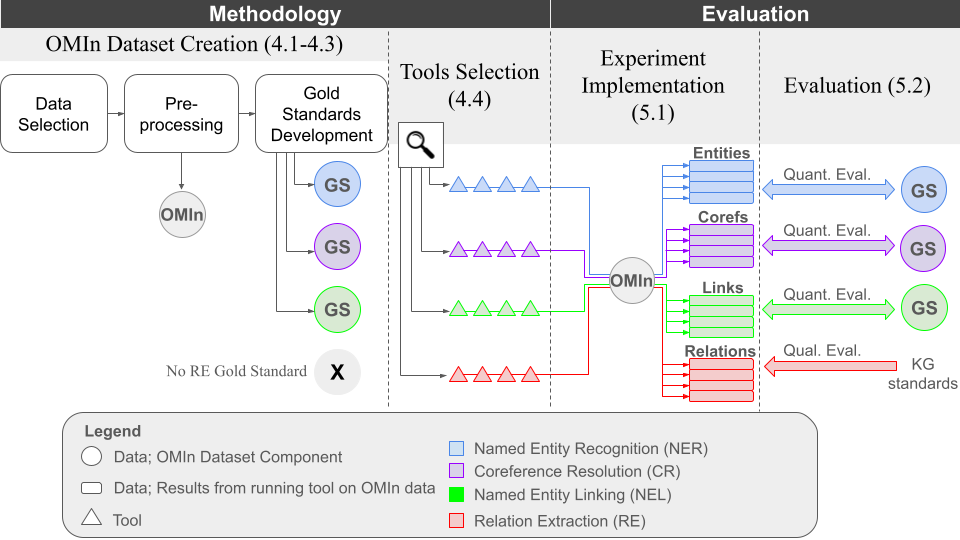}
    \caption{Conceptual Overview of Methodology and Evaluation Strategy Used in This Study. To create the OMIn dataset, we proceed through data selection (Subsection \ref{sec:data_selection}), pre-processing (Subsection \ref{sec:FAA_Data}), and Gold Standards (GSs) development (Subsection \ref{sec:GS}). Then, we select four tools for each stage of the KE workflow (Subsection \ref{sec:tools_selection}) These sixteen resultant tools are then implemented on the OMIn dataset, each creating a set of results in the form of named entities (NER), co-references (CR), linked entities (NEL), or relational triples (RE). The experimental setup for this implementation is captured in Subsection \ref{sec:experimental_setup}. These results are then evaluated against their respective GSs, or in the case of RE, against qualitative standards for knowledge representation in KGs. Development and implementation of evaluation metrics is discussed in Subsection \ref{sec:eval_metrics}.}
    \label{fig:Methodology_Workflow}
\end{figure}

To evaluate KE approaches in the maintenance domain, we construct a benchmark dataset and systematically assess tool performance across four key tasks: NER, CR, NEL, and RE. Figure~\ref{fig:Methodology_Workflow} illustrates the end-to-end workflow, which begins with the collection and curation of domain-specific data and proceeds through gold standard development and tools selection, and on to implementation and evaluation. Each step is detailed in the subsections that follow or Section \ref{sec:Eval}, Evaluation, as indicated in the diagram.

\begin{figure}[ht!]
    \centering
    \includegraphics[width=\textwidth]{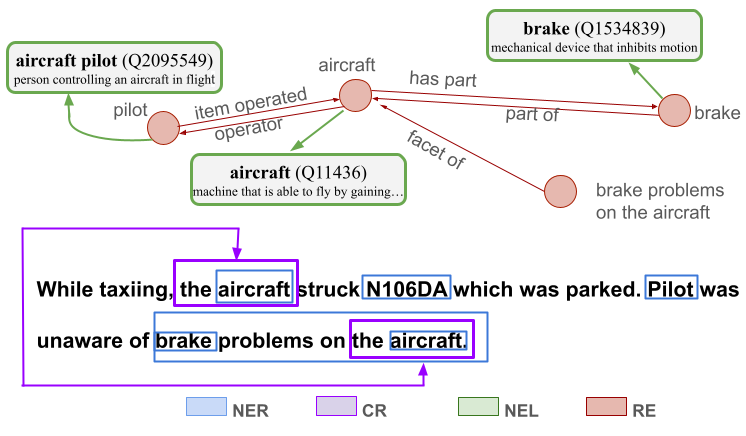}
    \caption{Sample KE Tasks Implementation.
    Here, the KE Tasks are applied to the sentence on the bottom from OMIn to generate the entities and coreferences annotated on the sentence itself as well as the relational triples and links represented in the graph on the top. Named entities, like \textit{aircraft} and \textit{Pilot}, are denoted in blue, and may be understood as either direct results from an NER implementation or intermediate results from an NER subtask via NEL and RE. Then, the CR system (purple) recognizes that the \textit{aircraft} refers to the same entity in different parts of the sentence, ensuring information relating to the aircraft is consolidated around one node. NEL (green) connects recognized entities to their corresponding Wikidata entries, such as \textit{aircraft} (Q11436) and \textit{aircraft pilot} (Q2095549). Finally, RE (red) identifies relationships between entities, with the red edges representing Wikidata properties, such as the pilot \textit{operating} the aircraft or the brake being a \textit{part of} the aircraft.}
    \label{fig:KE_Tasks_OMIn_Example}
\end{figure}

The outputs of each KE task, entities, co-references, links, and relational triples, capture complementary aspects of knowledge within each OMIn record. Figure ~\ref{fig:KE_Tasks_OMIn_Example} illustrates how the extracted results interact with the original data.

\subsection{Data Selection}
\label{sec:data_selection}

While there are few open-source maintenance and operations datasets, we want to highlight four noteworthy sources.
\begin{itemize}

\item NASA's Prognostics Center of Excellence (PCOE) offers several datasets that track the performance, operating conditions, and indications of damage of components such as bearings and batteries, as well as machines such as millers. However, there is no free-response natural language text in these records (\cite{nasa_pcoe}).

\item NASA's Aviation Safety Reporting System (ASRS) maintains a database of aviation safety incident and situation reports. It features a search engine that enables users to filter and download data. Additionally, ASRS provides PDF files containing a small selection of 50 records pertinent to 30 different topics (\cite{nasa_asrs_dataset}).

\item MaintNet's aviation dataset from the University of North Dakota Aviation Program (2012-2017) has 6,169 records. The records have ``Problem" and ``Action" fields and describe how problems were fixed. Although the dataset is fairly large, many items are repetitive, brief, and lack narrative context (\cite{akhbardeh2020maintnet}).
 
\item The Federal Aviation Administration (FAA) maintains a database of aviation accident and incident reports spanning more than 45 years. The Accident \& Incident Data (AID) dataset contains reports with a description of each incident along with details such as airplane type, an accident-type code, and more (\cite{faa_accident_incident_dataset}).
\end{itemize}

After reviewing the available options, we decided to use the FAA's AID dataset for our evaluation. While ASRS offers similar data, it includes numerous redacted proper nouns that could potentially confuse the KE tools. Although MaintNet's ``Problem" and ``Action" fields could be useful for training a model to aid in problem-solving within the maintenance domain, they lack sufficient narrative and context for effective KE evaluation.

We recognize that AID reports only describe events noticed by the pilot and ground crew during flight operations, not actions taken during direct maintenance. Although we use a subset of the AID dataset consisting of maintenance-related accidents and incidents, the resultant dataset is still distinct from a maintenance dataset made up of logs written by a maintenance technician. However, AID is a valuable starting point for operational and maintenance KE in several key ways: its short document size, frequent use of domain-specific shorthand and acronyms, and use of identification codes for vehicles and system components.

Building on the related work, our study evaluates KE tools using a novel dataset tailored to the maintenance and aviation domains. The following sections introduce this dataset and outline our evaluation methodology, providing insights into the challenges and opportunities for advancing domain-specific KE.

\subsection{Dataset Creation: OMIn}
\label{sec:FAA_Data}
\begin{figure}[ht!]
    \centering
    \includegraphics[scale=0.5]{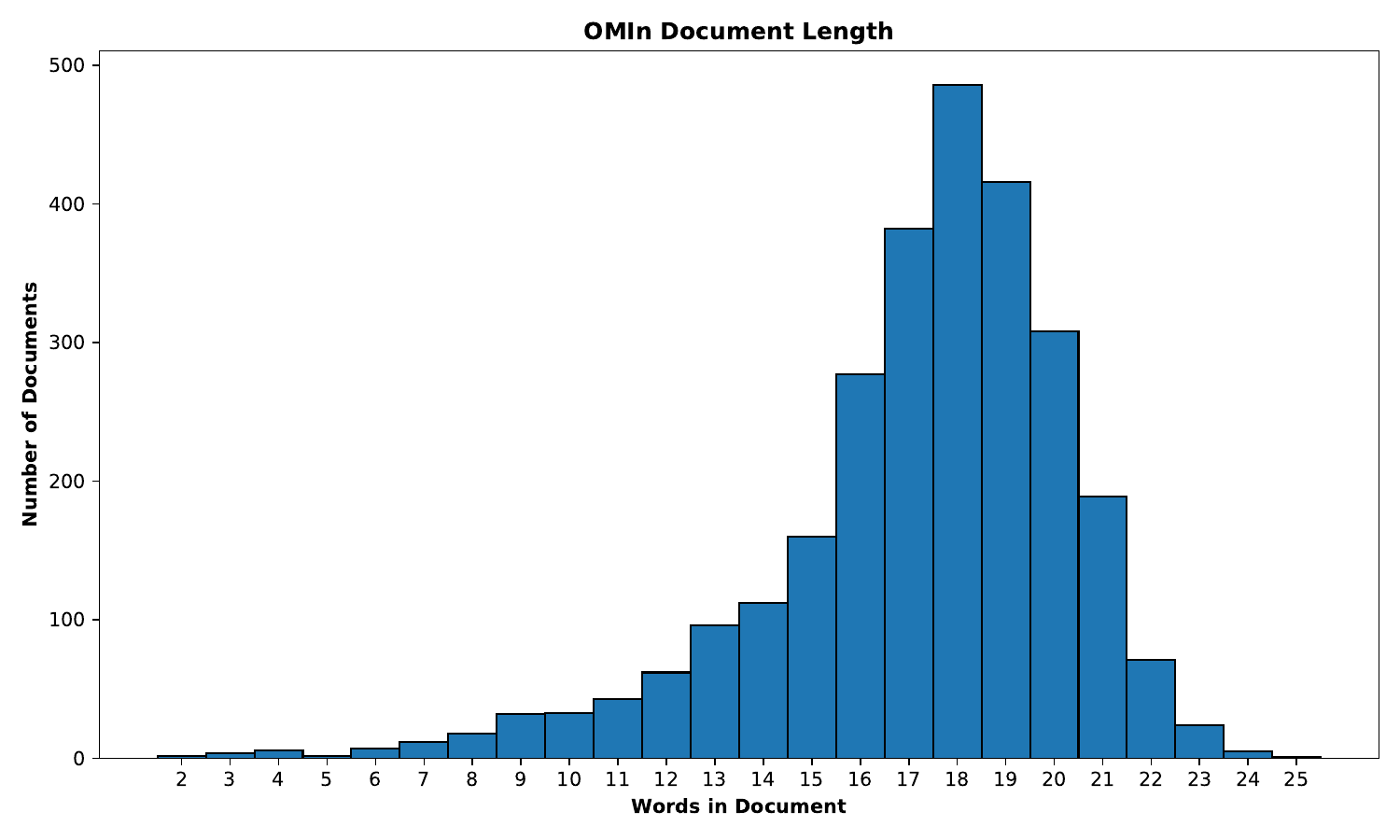}
    \caption{Distribution of Document Lengths in OMIn. The OMIn Dataset features 2748 short documents, usually 1-3 incomplete sentences, which are drawn from accident/incident reports captured in AID. The documents range between 2 and 25 words. The mean is 17.23, and the standard deviation is 3.14. The Q1 is 16; the median is 18; and Q3 is 19.}
    \label{fig:words_per_doc}
\end{figure}

\begin{figure}[ht!]
    \centering
    \includegraphics[width=\textwidth]{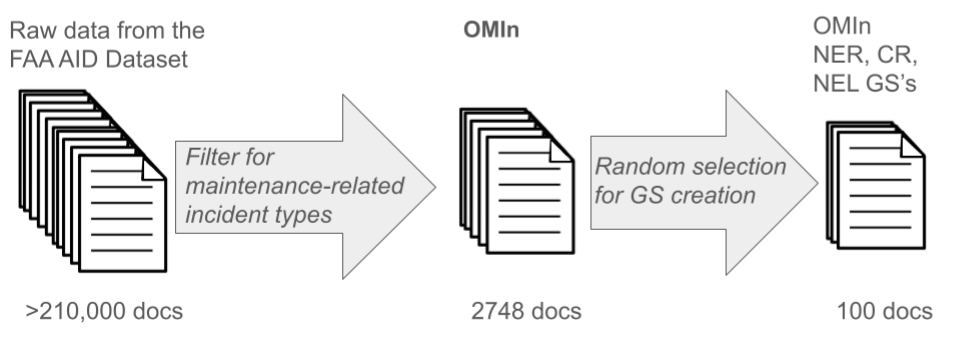}
    \caption{OMIn Dataset Curation. The OMIn Dataset is a subsection of maintenance-related incidents from the FAA Accident/Incident Dataset. A random selection of 100 records from OMIn were chosen as the basis for task-specific gold standards. The remaining 2648 documents in OMIn are un-labeled.}
    \label{fig:dataset_processing}
\end{figure}

We downloaded all available records from the FAA AID dataset spanning from 1975 to 2022 in June 2022, which totaled in excess of 210,000 records.\footnote{The dataset is updated regularly, and at the time of download, the latest record was from May 24, 2022.} We examined the fields of the records, as detailed in our data documentation, and found that 8 of the 116 incident types were related to maintenance. We selected records belonging to the 8 maintenance-related incident types and excluded those without textual description fields, resulting in a refined dataset of 2,748 records. We did not perform spellchecking or acronym resolution since our initial objective was to establish a baseline using the raw data. We refer to our subset of AID as the Operations and Maintenance Intelligence (OMIn) dataset. We treat each record in the OMIn dataset as a separate document when loading the dataset into the systems we evaluate. Figure \ref{fig:words_per_doc} shows the distribution of document lengths in OMIn. Figure \ref{fig:dataset_processing} illustrates how the OMIn dataset is curated from AID.

Some of the tools allowed the OMIn dataset to be passed in directly as plain text. However, other tools were more challenging to run and required the FAA Data to be preprocessed. Table \ref{tab:FAA_Process_Direct} shows which tools allowed data to be passed directly as plain text. The tools trained on CoNLL-2012, which were ASP and s2e-coref, required data in an annotated CoNLL-2012 file format rather than simple strings. Therefore, we adapted the FAA data to the CoNLL-2012 format, which organizes documents in a tabular structure with each word on a separate line and more than 11 columns detailing the word's semantic role. See \ref{sec:CoNLL-2012_Pre-Processing} for more information on our process.
 \begin{table}[ht!]
    \centering
    \footnotesize
        \begin{tabular}{|c|c|c|c|c|c|c|c|} \hline
         \multicolumn{2}{|c|}{\textbf{NER}}&  \multicolumn{2}{|c|}{\textbf{Coref}}&  \multicolumn{2}{|c|}{\textbf{NEL}}&  \multicolumn{2}{|c|}{\textbf{RE}}\\ \hline 
         spaCy&  ✓&  ASP&  ✘&  BLINK&  ✓&  REBEL& ✓\\ \hline 
         flair&  ✓&  coref-mt5&  ✘&  spaCy Entity Linker&  ✓&  UniRel& ✓\\ \hline 
         stanza&  ✓&  s2e-coref&  ✘&  GENRE&  ✓&  DeepStruct& ✘\\ \hline 
         nltk& ✓&  neuralcoref&  ✓&  ReFinED&  ✓&  PL-Marker& ✘\\ \hline
    \end{tabular}
    \caption{Tools That Can Directly Process OMIn Text.}
    \label{tab:FAA_Process_Direct}
\end{table}

\subsection{Gold Standard Development}
\label{sec:GS}
Gold Standards (GSs) for each KE task can vary depending on the criteria selected by the research team responsible for their creation. Our team began by considering how to structure a maintenance ontology. From there, we tailored the gold standards to align with and augment the maintenance ontology.
Due to the significant manual effort this process required, we selected a sample of 100 records from our dataset to create the gold standards.  We intend to grow the number of curated records with community collaboration.

\begin{tcolorbox}[colback=blue!5!white, colframe=blue!75!black, title=Gold Standard vs Ground Truth]
We assign the term gold standard versus ground truth as the dataset was human-curated and represents language relations and classifications that are not universally defined nor understood as ground truth measures.
\end{tcolorbox}

\subsubsection{Named Entity Recognition Gold Standard}
\label{sec:NER_GS}
There are several open-source standards for NER annotation, used in shared tasks such as CoNLL-2003, ACE-2005, and CoNLL-2012. Each of these shared tasks has developed its own set of annotation guidelines and set of named entity types. The entity types used in the CoNLL-2012 NER task correspond to the schema defined in the OntoNotes 5.0 corpus, and the 18 entity types are most commonly referred to as the OntoNotes 5.0 set. We adopt that convention in this paper. Most NER systems classify entities that fall into ENAMEX types such as person, location, and organization, with some also classifying time and date (TIMEX) types and number (NUMEX) types. No NER benchmark focuses on entity types relevant to data from the maintenance or aviation domains. Absent of these types, we gathered a small team of trained annotators who labeled un-typed named entities that constituted the essential information in each record of the FAA dataset. We started by following guidelines for ACE-2005, and added TIMEX and NUMEX entities according to OntoNotes 5.0. We then added additional entities for essential aviation maintenance information, including aircraft parts, systems, phases and types of operations, and equipment failures. 
We refer to this GS as the Un-Typed FAA (UTFAA) GS and record our annotation guidelines in \ref{sec:NER_Annotation_Guidelines}.

For example, in the record, ``Narrative: The cargo door was latched before takeoff by Mr. Bowen. Runway conditions at Steven's Village was extrem," the entities we annotate are ``cargo door", ``takeoff", ``Mr. Bowen", ``runway conditions", and ``Steven's Village."

We also annotated the FAA dataset following the guidelines used in CoNLL-2003, ACE Phase 1, ACE-2005, and OntoNotes 5.0. We refer to these as the benchmark-annotated GSs when grouped collectively or as CoNLLFAA, ACE1FAA, ACE05FAA, and ONFAA, respectively. ACE1FAA is equivalent to ACE05FAA with the vehicle-type entities removed, which leaves the set of entities which NLTK is trained to recognize. Meanwhile, PL-Marker, whose NER subtask we evaluate, recognizes the set of entities in ACE-2005. A summary of the gold standard entities and their distribution across entity types can be seen in Table \ref{tab:NER_GS_Entity_Types}. Note that the different gold standards may tag entities differently, even for the same entity type. This is due to differences in annotation guidelines across the benchmarks.

We compared the three resulting sets of named entities against our set of named entities and the overlap as recorded in Table \ref{tab:NER_GS_Agreement}. Total refers to the total number of entities generated for the set of 100 sample records by each benchmark-annotated GS. Match and Partial Match refer to the number of entities in each benchmark-annotated GS that match or partially match an entity in our GS, regardless of label. Overlap is the sum of the matches and partial matches divided by the total number of entities in our GS, which is 510. For evaluation, we use both our un-typed gold standard and the benchmark-annotated gold standards.

\begin{table}
    \centering
    \begin{adjustbox}{max width=\textwidth}
    \begin{tabular}{|c|c|c|c|c|c|} \hline 
         &  \textbf{CoNLLFAA}&  \textbf{ACE05FAA}&\textbf{ACE1FAA}&  \textbf{ONFAA}&\textbf{UTFAA}\\ \hline
 \textbf{Total}& \textbf{44}& \textbf{195}& \textbf{122}& \textbf{61}&\textbf{509}\\ \hline
 PER& 3&  49&49& 3&--\\ \hline 
 ORG& 11&  11&11& 11&--\\ \hline 
 LOC& 21&  15&15& 0&--\\ \hline 
 MISC& 9&  --&--& --&--\\ \hline 
 GPE& --&  14&14& 14&--\\ \hline 
         FAC&  --&   33&33&  7&--\\ \hline 
 VEHICLE& --& 73& --& --&--\\ \hline 
         PRODUCT&  --&   --&--&  9&--\\ \hline 
         QUANTITY&  --&   --&--&  6&--\\ \hline 
 CARDINAL& --& --& --& 3&--\\ \hline 
 DATE& --& --& --& 4&--\\ \hline 
 TIME& --& --& --& 4&--\\ \hline
    \end{tabular}
    \end{adjustbox}
    \caption{Distribution of Entity Types across NER Gold Standards. The ``--" symbol indicates that the Entity Type does not belong to the corresponding standard.}
    \label{tab:NER_GS_Entity_Types}
\end{table}

\begin{table}[ht!]
    \centering
    \footnotesize
    \begin{tabular}{|l|c|c|c|l|} \hline
          &\textbf{Total}&\textbf{Match}&\textbf{Partial Match}&\textbf{Overlap}\\ \hline 
            CoNLLFAA&44&36& 8&0.086\\ \hline 
          ACE05FAA&195&133&54&0.35\\ \hline
          ACE1FAA&122&89&26&0.22\\ \hline
          ONFAA&61&52&9&0.12\\\hline
    \end{tabular}
    \caption{Agreement of NER on Gold Standard}
    \label{tab:NER_GS_Agreement}
\end{table}

\subsubsection{Coreference Resolution Gold Standard}
\label{sec:CR_GS}
There were no maintenance-specific adjustments necessary for CR, so we adhered to CoNLL-2012 guidelines and the OntoNotes 5.0 phrase tagging guidelines (\cite{pradhan_conll-2012_2012}). In CoNLL-2012, co-referential entities may include a broad scope of phrases with potentially differing grammatical structures and roles, linked by a common reference to the same real-world entity.

\subsubsection{Named Entity Linking Gold Standard}
\label{sec:NEL_GS}
Our NEL gold standard is based on the named entities identified in our UTFAA gold standard. We found Wikidata Q-identifiers (QIDs) by manually looking up each entity and listing the most specific Q-identifier if there was a correct one. All of the NEL tools we evaluate use QIDs or Wikipedia-based identifiers, such as titles or links to entries, which can be easily translated into QIDs. This enabled us to directly compare the links predicted by each NEL tool with those in the gold standard.

We also created a Flexible NEL GS, which includes additional entity-QID links, motivated by the fact that differing mention spans may change the appropriate QID for each entity. For example, in the sentence ``While taxiing lost nosewheel steering and brakes", we have ``nosewheel steering" as an entity in our UTFAA NER GS. If an NEL tool only recognizes ``steering" as the entity and links it to the QID for steering correctly (Q18891017), this would be excluded from the evaluation set under strong entity-matching and counted as incorrect under weak entity-matching. Our Flexible GS makes a flexible evaluation possible, where if an exact match for ``nosewheel steering" is not found, the evaluator moves to a secondary entity-QID link, (``steering", Q18891017), and evaluates the predicted entity-QID against it. To accomplish this, we included primary, secondary, and up to tertiary entity-QID pairs for entities such as ``nosewheel steering," where sub-spans of the primary entity share the same semantic role in the sentence.

\subsubsection{Absence of Relation Extraction Gold Standard}
\label{sec:RE_GS}
Unlike the other KE tasks considered in this study, we do not provide a gold standard (GS) for RE. The primary impediment to establishing such a GS is the heterogeneity of the relations that different tools recognize. Each evaluated tool is trained on a distinct set of relationships which Table \ref{tab:RE_Relations_Sets} highlight showing illustrative training data relationships for each RE model, as well as a subset of the relation sets they employ. Although some conceptual overlap exists, no unified ontology currently aligns these disparate sets into a coherent framework that could serve as the basis for a universal gold standard.

\begin{table}
    \centering
    \begin{adjustbox}{max width=\textwidth}
    \begin{tabular}{|c|c|c|c|} \hline 
         \textbf{Training Data}&  \textbf{Tool}&  \textbf{\# Relations}&  \textbf{Sample Relations}\\ \hline 
         Wikidata&  REBEL&  220&  has part, part of, has effect, has cause, location, subclass of, ...\\ \hline 
 NYT& UniRel, DeepStruct& 25&location/contains, person/company, company/founders, ...\\ \hline 
 ACE05& PL-Marker& 6&PER-SOC, ART, PHYS, ORG-AFF, ...\\ \hline 
         SciERC&  PL-Marker&  7&  PART-OF, USED-FOR, FEATURE-OF, CONJUNCTION, ...\\ \hline
    \end{tabular}
    \end{adjustbox}
    \caption{RE Tool Models Predict Different Relations Depending on Training Data}
    \label{tab:RE_Relations_Sets}
\end{table}

In principle, one could construct a specialized maintenance ontology that enumerates domain-relevant relations and subsequently benchmark each tool against it. However, such an approach introduces several methodological complications. Relations often differ substantially in their level of granularity and intended scope across datasets. For example, ACE-2005 employs a high-level “PART-WHOLE” relation that encompasses both geographic and non-geographic entities, while the New York Times (NYT) corpus relies on a narrower “location contains” relationship. In contrast, Wikidata’s ontology includes multiple properties—such as “located in the administrative territorial entity,” “part of,” and “location”—that articulate spatial and hierarchical relations at different levels of specificity. Selecting any one level of granularity for a putative gold standard would inevitably privilege certain tools’ relational inventories and disadvantage others. Moreover, introducing relations absent from a given tool’s training data would yield uniformly poor performance on those relations, thereby obscuring meaningful comparisons across tools.

The only rigorous way to ensure fairness and consistency would be to fine-tune all RE models on a common, domain-specific ontology. Yet, constructing such an ontology and annotating a sufficiently large corpus of maintenance text represents a significant investment of time and effort. Furthermore, expert input from the maintenance community would be essential for achieving conceptual clarity and operational relevance. We thus consider this endeavor more suitably reserved for future collaborative work rather than as part of the present study.

Consequently, we do not report an F1 score against a unified GS for RE. Instead, as described in Section \ref{sec:RE_Eval}, we rely on accuracy and other metrics that do not presuppose a single ontology. We anticipate that subsequent studies, supported by broader community engagement, will produce a maintenance-specific ontology and corresponding gold standard that enable robust, equitable evaluations of RE performance.

\subsection{Tools Selection}
\label{sec:tools_selection}
We selected sixteen tools for evaluation, with four dedicated to each KE task. The primary criterion guiding this selection was the availability of off-the-shelf, open-source solutions specifically fine-tuned and benchmarked for NER, CR, NEL, and RE. This approach enables us to compare the tools’ expected in-domain performance with their observed zero-shot results on our maintenance dataset, thereby providing a stable and interpretable baseline.

Although it is true that the latest LLMs, including transformer-based systems such as GPT variants or LLaMA, have demonstrated state-of-the-art performance in various NLP tasks, their general-purpose capabilities often come at the cost of increased complexity in domain-specific applications. Adapting these models for KE tasks currently involves additional steps, such as prompt engineering, specialized prompting frameworks, and robust guardrails to mitigate unpredictable outputs. Although these LLMs excel at open-ended reasoning and question-answering, they lack the immediate, task-specific fine-tuning that would allow for straightforward evaluation in our zero-shot setting.

Our decision to focus on models already proven in KE benchmarks, many of which have transparent code bases and straightforward deployment procedures, aligns with our emphasis on trusted and reproducible AI solutions. It also establishes a clear starting point. By first understanding how specialized KE tools perform without adaptation, we create a benchmark against which future efforts, such as applying domain adaptation, fine-tuning, or integrating cutting-edge LLMs, can be measured. Thus, our findings lay the groundwork for subsequent experimentation with advanced transformer-based models, once their methods for reliably performing KE tasks are more thoroughly developed.

Additionally, we excluded models exceeding 13B parameters, as such large architectures typically require substantial computational resources, often facilitated through cloud-based services. Although state-of-the-art in many tasks, these models introduce practical concerns related to infrastructure, cost, and security, particularly for organizations handling sensitive and confidential data. By highlighting tools that can be run on-premises and without extensive cloud dependencies, we underscore our commitment to delivering trustworthy, secure solutions aligned with operational needs in the maintenance domain.

The website paperswithcode.com was a helpful resource for finding open-source models that have reported high scores on popular benchmarks (\cite{noauthor_papers_nodate}).

\subsubsection{Named Entity Recognition Tools}
Since there are a variety of high-performing NER tools available, we prioritized evaluating readily available tools over those with the highest performance. This led us to the well-known NLP toolkits spaCy (\citet{HonnibalMontani2020}), flair (\citet{akbik2018coling}), stanza (\cite{shan_english_2023}), and NLTK. Table \ref{tab:NER_Tools} summarizes the NER tools selected.

\begin{table}
    \begin{adjustbox}{max width=\textwidth}
    \centering
    \begin{tabular}{|c|c|c|} \hline 
         Tool&  Entity Set&  Benchmark Performance (F1)\\ \hline 
         spaCy EntityRecognizer&  OntoNotes 5.0&  Benchmark Not found\\ \hline 
         Flair NER&  CoNLL-2003, Ontonotes 5.0&  94.09 (CoNLL-03), 90.93 (OntoNotes)\\ \hline 
         Stanza NERProcessor&  CoNLL-2003, Ontonotes 5.0&  92.1 (CoNLL-03), 88.8 (OntoNotes)\\ \hline 
         NLTK ne\_chunk&  ACE Phase 1 + GSP\footnote{NLTK ne\_chunk recognizes the five entity types found in ACE Phase 1 (\cite{entity_detection_2002}), as well as a sixth type, Geographical-Social-Political Entity (GSP), an early form of Geo-Political Entity (GPE), found in ACE-Pilot (\cite{entity_detection_2000}).}&  No Benchmark Found\\ \hline
    \end{tabular}
    \end{adjustbox}
    \caption{NER Tools Selected for Evaluation}
    \label{tab:NER_Tools}
\end{table}

\subsubsection{Coreference Resolution Tools}
Many coreference resolution tools have been bench-marked on well-recognized shared tasks and challenges such as CoNLL-2012 (\cite{pradhan_conll-2012_2012}), Gendered Ambiguous Pronouns (GAP) (\cite{webster_gendered_2019}), and the Winograd Schema Challenge (WSC) (\cite{levesque_winograd_2012}). GAP and WSC evaluate a model's ability to link pronouns to their antecedents. However, CoNLL-2012 includes noun phrases that refer to one another, such as ``vehicles" and ``armored vehicles." Since maintenance data rarely includes pronouns and often refers to system components in multiple ways, we decided that tools trained for CoNLL-2012 would be the best fit to evaluate maintenance data. The tools selected for this evaluation are summarized below.

\begin{itemize}

\item \textit{Autoregressive Structured Prediction with Language Models (ASP)} is a T5-based model developed by Google Research in 2022 (\cite{liuAutoregressiveStructuredPrediction2022}). ASP proposes a conditional language model trained over structure-building actions instead of strings. This enables it to capture the structure of the sentence more effectively and build the target structure step by step. It performs NER, CR, and RE. Its CR models include three based on different sizes of flant5 (base, large, and xl), and one based on T0-3B. It achieved an F1 score of 82.3 for CR n on CoNLL-2012.

\item \textit{Coref\_mt5} is a mT5 model developed by Google Research in the work \textit{``Coreference Resolution through a seq2seq Transition-Based System"} (~\cite{bohnetCoreferenceResolutionSeq2seq2022}).
This model leverages the mT5 architecture and employs a seq2seq approach. It encodes a single sentence along with its preceding context as input and generates an output with predicted coreference links. The model utilizes a transition-based system to extend discovered coreference chains and establish new ones in subsequent sentences. It achieved an F1 score of 83.3 on CoNLL-2012.

\item \textit{Start-To-End Coreference Resolution (s2e-coref)} introduces a lightweight approach that avoids constructing span representations \cite{kirstainCoreferenceResolutionSpan2021}. Instead, it uses contextualized representations of the boundaries of spans to score the likelihood of a coreference between a mention and potential antecedents. It is based on longformer-large and achieved an F1 score of 80.3 on CoNLL-2012.

\item \textit{NeuralCoref} is a coreference resolution pipeline component developed by the spaCy team. It uses the spaCy parser for mention-detection and ranks possible mention-coreference pairs using a feedforward neural network developed by Clark and Manning, Stanford University. The Clark and Manning network achieved an F1 score of 74.23 on CoNLL-2012 in 2016.

\end{itemize}

\subsubsection{Named Entity Linking Tools}
We chose three NEL tools that have achieved state-of-the-art results and are compatible with Wikidata. Additionally, we decided to evaluate the spaCy EntityLinker because of spaCy's wide recognition.

\begin{itemize}

\item \textit{Better Entity LINKing (BLINK)} introduces a two-stage zero-shot entity linking algorithm, with a bi-encoder for dense entity retrieval and a cross-encoder for re-ranking (\cite{wuScalableZeroshotEntity2020}). It uses a predefined catalog of entities from Wikipedia and uses the first few sentences of their Wikipedia summaries as context. It uses BERT as a base model for its encoders. BLINK does not do NER on its own but utilizes flair. It achieved a 76.58\% accuracy on the Zero-shot EL dataset and a 94.5\% accuracy on TACKBP-2010.

\item \textit{spaCy EntityLinker} is spaCy's NEL pipeline component. It uses InMemoryLookupKB to match mentions with external entities. InMemoryLookupKB contains Candidate components that store basic information about their entities, like frequency in text and possible aliases.

\item \textit{Generative ENtity REtrieval (GENRE)} employs a BART-based seq2seq model to autoregressively generate entity identifiers (\cite{decaoAutoregressiveEntityRetrieval2021}). Additionally, GENRE uses a constrained decoding strategy that forces each generated identifier to be in a predefined candidate set, ensuring that the generated output is a valid entity name. It achieved micro-F1 scores ranging from 77.3 to 94.3 on both in-domain and out-of-domain benchmarks.

\item \textit{Representation and Fine-grained typing for Entity Disambiguation (ReFinED)} uses fine-grained entity types and entity descriptions to perform mention detection, fine-grained entity typing, and entity disambiguation in a single forward pass (\cite{ayoolaReFinEDEfficientZeroshotcapable2022a}). Similar to BLINK, ReFinED uses a catalog of entities from Wikipedia (and Wikidata) with context from their summaries, and new entities can be added to the catalog without retraining. ReFinED includes three NEL models, wikipedia\_model, wikipedia\_model\_with\_numbers, and aida\_model, all based on RoBERTa. It achieves micro-F1 scores ranging from 78.2 to 94.8 on both in-domain and out-of-domain benchmarks.

\end{itemize}

\subsubsection{Relation Extraction Tools}
We chose the following Relationship Extraction (RE) tools because they had reached state-of-the-art performance on widely recognized benchmarks, including CoNLL-2004, NYT, and ACE-2005. Note that each benchmark dataset uses a different set of relations, depending on the subject of the data. None of these sets of relations were adequate to capture all of the most relevant information for maintenance data. Note that reference to relational triples or triplets below refer to structures in the form (\textit{subject}, \textit{relation}, \textit{object}).

\begin{itemize}

\item \textit{Relation Extraction By End-to-end Language generation (REBEL)} uses an autoregressive seq2seq model based on BART to express relation triplets as a sequence of text (\citet{huguet-cabot-navigli-2021-rebel-relation}). It finds 220 relation types, a subset of Wikidata properties chosen by the REBEL team. REBEL achieves an F1 of 91.76 on NYT, 71.97 on CoNLL-2004, and 90.39 on Re-TACRED.

\item \textit{Unified Representation and Interaction for Joint Relational Triple Extraction (UniRel)} jointly encodes entities and relations and captures interdependencies between entity-entity interactions and entity-relation interactions through the proposed Interaction Map (\cite{tangUniRelUnifiedRepresentation2022}). It is based on bert-base-cased, and trained on the New York Times (NYT) dataset. It consists of 24 relation types. UniRel achieves an F1 of 93.7 on NYT and 94.7 on WebNLG.

%\item \textit{DeepStruct} was developed by \citet{wangDeepStructPretrainingLanguage2023} in 2022. DeepStruct is pretrained on several task-agnostic corpora to learn to generate structures from text. DeepStruct is provided as four pretrained models, trained on CoNLL-2004, ADE, NYT, and ACE-2005; each model finds the set of relation types associated with those datasets. We selected the NYT pretrained model has the highest RE F1 score of 84.6. Additionally, both UniRel and DeepStruct use the NYT dataset, which ensures consistency between tools.

\item \textit{DeepStruct} (~\citet{wangDeepStructPretrainingLanguage2023}) introduced a model pretrained on task-agnostic corpora to enhance language models' ability to generate structured outputs, such as entity-relation triples, from the text. DeepStruct is a unified model designed to generalize across a wide range of datasets and tasks, including RE, NER, and event extraction, among others.  However, some steps need to be followed to perform inference on a dataset that is not included in DeepStruct's list of pretrained datasets. The dataset schemes related to the RE task used in the pretrained model are from the datasets NYT, CoNLL04, ADE, and ACE2005. We selected the NYT setup to use in our evaluation, with the highest RE F1 score of 84.6. Additionally, both UniRel and DeepStruct use the NYT dataset, which improves consistency between tools.

\item \textit{Packed Levitated Marker (PL-Marker)} uses markers in the encoding phase to capture the interrelation between span pairs (\cite{ye2022plmarker}). The novelty of PL-Marker's approach is its neighborhood-oriented span-packing strategy. PL-marker provides NER and RE models trained on the SciERC and ACE-2005 datasets, as well as three other NER models. Their BERT-based models, which we evaluate, achieved a 69.0 F1 score on ACE-2005 and a 53.2 F1 score on SciERC in RE. We also evaluate their Albert-xxl-based model which achieved state-of-the-art on ACE-2005 RE with an F1 score of 73.0.

\end{itemize}

\section{Evaluation}
\label{sec:Eval}

Our evaluation process focuses on the tools' ability to handle unseen elements in a zero-shot scenario.  Zero-shot here refers to evaluating tools on the FAA dataset without prior fine-tuning or specific training in aviation or maintenance domains.

\subsection{Experimental Setup}
\label{sec:experimental_setup}

We followed the guidelines provided in each tool's documentation to set up and generate output. To reproduce our results, see our step-by-step methods documented in our repository \cite{our_repo}.

While some tools were straightforward to implement, others presented challenges. The rapid pace of advancements in NLP means that library dependencies quickly become outdated. There can also be a complicated environment resolution setup if the accompanying requirements files do not specify all necessary package versions. Other challenges included a lack of clear indications of the GPU requirements needed for model deployment, and occasional bugs. The greatest challenge was the lack of clear documentation on how to use these tools with our data. This required us to carefully review the tool code base to understand the required data format and identify any necessary adjustments for our dataset compared to the benchmark data. In our GitHub documentation for each tool, we offer a ``Reproducibility Rating" along with an account of the challenges we encountered during setup.

For tools that ship multiple models, we strove to evaluate the models most applicable to our task. For spaCy tools, we selected the small and large models. For RE tools, we select the best-performing model as well as others we believe would be informative to evaluate. All models are trained on English data. Table \ref{tab:models} summarizes the models implemented for each tool that has more than one model available.

\begin{table}[ht!]
\centering
\renewcommand{\arraystretch}{1.4}
\footnotesize
\begin{adjustbox}{max width=\textwidth}
\begin{tabular}{|C{2.5cm}|C{6cm}|C{6cm}|}\hline
\textbf{Tool} & \textbf{Models Implemented} & \textbf{Models Not Implemented} \\ \hline
flair & CoNLL-03, OntoNotes & - \\ \hline
spaCy EntityRec & en\_core\_web\_sm, en\_core\_web\_lg & en\_core\_web\_md, en\_core\_web\_trf \\ \hline 
stanza & conll03\_charlm, all ontonotes and ontonotes-ww-multi models& conll03\_electra-lg, all non-conll03 \& ontonotes\\ \hline
ASP & flant5\_base, flant5\_large, flant5\_xl, t0\_3b& all non-CR models\\ \hline
neuralcoref & en\_core\_web\_sm, en\_core\_web\_lg& en\_core\_web\_md, en\_core\_web\_trf\\  \hline
BLINK & biencoder, crossencoder & - \\ \hline
spaCy EntityLink & en\_core\_web\_sm, en\_core\_web\_lg& en\_core\_web\_md, en\_core\_web\_trf\\ \hline
GENRE & BART\_base E2E EL & - \\ \hline
ReFinED & wikipedia\_model\_with\_numbers, wikipedia\_model, aida\_model& -\\ \hline
UniRel & NYT & WebNLG \\ \hline
PL-Marker & scibert-uncased (SciERC), bert (ACE-05), albert-xxl (ACE-05) & NER models w/o RE \\ \hline
%DeepStruct & NYT JER& ACE-2005 JER, CoNLL-04 JER, non-JER models\\ \hline
\end{tabular}
\end{adjustbox}
\caption{Models Implemented For Each Tool}
\label{tab:models}
\end{table}

REBEL provided one English-only model, rebel-large. However, REBEL allows this model to be used in several ways. Rebel-large can be downloaded directly and implemented with user-visible hyper-parameters, or loaded as a component in either a transformers or spaCy pipeline. We chose to test the direct implementation of the model as well as the transformers pipeline.  In our testing, the transformers pipeline produced consistent results, while the direct implementation did not, even with the same hyper-parameters.
%
%However, during testing, we found that the transformers pipeline implementation was deterministic, but the direct implementation was not, regardless of the hyper-parameters selected. 
%
%
%We chose not to evaluate the results from the multiple runs and different hyper-parameters of the direct implementation of REBEL on the FAA dataset, because we evaluate RE by hand. However, the results from these runs, including those with varying hyper-parameters, are still available on our GitHub repository for reference but without an accompanying manual evaluation.
%
%Since our evaluation of RE tasks is qualitative and requires extensive manual work, we chose not to assess the varied results from the direct implementation on the FAA dataset. However, these results, including those with different hyper-parameters, are available in our GitHub repository, albeit without manual evaluation.

For the Coref\_mt5, a T5X checkpoint from the top-performing mT5 model (\cite{bohnetCoreferenceResolutionSeq2seq2022}) was made publicly available\footnote{\href{https://github.com/google-research/google-research/tree/master/coref_mt5}{https://github.com/google-research/google-research/tree/master/coref\_mt5}}. However, we encountered issues when attempting to load this checkpoint. To address this, we utilized a PyTorch-converted version of the model, which was released by Ian Porada on HuggingFace.\footnote{\href{https://huggingface.co/ianporada/link-append-xxl}{https://huggingface.co/ianporada/link-append-xxl}}

DeepStruct required some additional steps to prepare it for inference on unseen datasets. First, the unseen dataset’s schema must be aligned with the one DeepStruct was trained on to ensure compatibility with the model’s recognized entity and relation types. Additionally, configuration files might need adjustments to properly format the input data. Considering the RE task, DeepStruct schemas are based on datasets like NYT, CoNLL-04, ADE, and ACE-2005. We chose the NYT schema for our evaluation, which produced the highest RE F1 score of 84.6. This also ensures consistency and fairness when comparing DeepStruct with UniRel.

\subsection{Evaluation Metrics}
\label{sec:eval_metrics}

We quantitatively evaluated NER, CR, and NEL tools by comparing their outputs to our established gold standard. In contrast, due to the lack of a gold standard for RE tools in the FAA dataset, we employed manual evaluation by a domain expert.

\subsubsection{Named Entity Recognition Evaluation}
\label{sec:NER_Eval}

We evaluated each NER tool against the un-typed UTFAA and against benchmark-annotated gold standards aligned with its training data (e.g., CoNLLFAA for models trained on CoNLL-2003). The key difference is that benchmark GS evaluation considers both entity types and spans, whereas UTFAA evaluation considers only spans and ignores types.

\begin{itemize}
    \item \textbf{Strong/Weak (UTFAA)}: F1 is computed on entity spans only. Strong matching requires exact span agreement; weak matching allows any substring overlap. For example, the gold entity \textit{Mr. Bowen} is scored incorrect under strong matching if a system outputs only \textit{Bowen}, but correct under weak matching.
    
    \item \textbf{Strict/Type/Exact/Partial (benchmark GS)}: Following SemEval \cite{segura2013semeval}, F1 metrics consider both entity spans and types. Strict and Type require correct entity types; Exact and Partial are label-agnostic. Exact/Partial correspond directly to strong/weak matching on UTFAA.
\end{itemize}

We used the work of \cite{batista_named-entity_2018} to implement both NER evaluations.

\subsubsection{Coreference Resolution Evaluation}
\label{sec:CR_Eval}
For CR, we report four standard metrics, which is important given the small size of maintenance datasets: our FAA CR Gold Standard contains only 18 coreferences. Using multiple metrics ensures robust evaluation.

\begin{itemize}
    \item \textbf{MUC} (\cite{vilain1995model}): measures link-based correctness.
    \item \textbf{B-CUBED} (\cite{bagga1998algorithms}): evaluates the accuracy of individual elements within clusters.
    \item \textbf{LEA} (\cite{moosavi2016coreference}): computes cluster-level precision and recall based on the similarity between predicted and true clusters.
    \item \textbf{CEAF} (\cite{luo2005coreference}): evaluates alignment of entities and links, considering the impact of each decision within clusters.
    \item \textbf{CoNLL-2012} (\cite{pradhan_conll-2012_2012}): unweighted average of MUC, B-CUBED, and CEAF, following CoNLL-2012.
\end{itemize}

We used \textit{corefeval} toolkit\footnote{\href{https://github.com/tollefj/coreference-eval}{https://github.com/tollefj/coreference-eval}} to implement all CR evaluations.

\subsubsection{Named Entity Linking Evaluation}
\label{sec:NEL_Eval}

We evaluated NEL tools using two complementary approaches: standard F1 score and Ontology-based Topological (OT) metrics derived from the structure of the underlying ontology. We also use three matching strategies, applied to both sets of metrics. Full computation details and worked examples, including Jiang–Conrath (JC) and class similarity, are provided in ~\ref{sec:NEL_Eval_Appendix}.

\begin{itemize}
    \item \textbf{Metrics}:
    \begin{itemize}
        \item \textbf{F1}: standard precision/recall-based F1, following \cite{shen2021entity, usbeck2015gerbil}.
        \item \textbf{Ontology-based Topological (OT) metrics}: computed over Wikidata using the KGTK Semantic Similarity system \cite{kgtk_similarity}, including Jiang–Conrath (JC) and class similarity scores.
    \end{itemize}

    \item \textbf{Matching strategies}:
    \begin{itemize}
        \item \textbf{Strong}: predicted entities must exactly match a gold entity.
        \item \textbf{Weak}: predicted and gold entities share any substring (see Section~\ref{sec:NER_Eval}).
        \item \textbf{Flexible}: evaluates predicted entities that strongly match entities in the Flexible NEL GS (see Section~\ref{sec:NEL_GS}).
    \end{itemize}
\end{itemize}

\subsubsection{Relation Extraction Evaluation}
\label{sec:RE_Eval}
We evaluate the RE output qualitatively on the same set of 100 records used to create gold standards for the other KE tasks. The relations extracted by an RE system may take several structures, including relational tables, XML files, and relational triples. For our evaluation, we use relational triples, since they are compatible with KG construction technologies such as Neo4j. To assess how well the generated triples support accurate KG construction, we use three KG evaluation metrics from Chapter 7 of \cite{Hogan_2021}, along with the total number of triples generated and the percentage of records containing triples. Our implementation methodology is detailed in ~\ref{sec:RE_Eval_Guidelines}.

\begin{itemize}
    \item \textbf{Syntactic accuracy}: measures grammatical agreement of triples.
    \item \textbf{Semantic accuracy}: measures correctness of meaning.
    \item \textbf{Consistency}: measures absence of contradictions across triples.
    \item \textbf{Number of Triples}: total number of triples generated.
    \item \textbf{Percent Docs with Triple}: percentage of records with one or more generated triples.
\end{itemize}

\section{Results}

\subsection{Overview}
As described in  Section \ref{sec:Eval}, we have quantitative evaluations for NER, CR, and NEL as well as qualitative scores for RE. Table \ref{tab:Overview_Results} shows the main results.\footnote{All items in parentheses following the name of the tool are model names unless otherwise stated. For ReFinED, wiki\_w\_nums is an abbreviation for the model, wikipedia\_model\_with\_numbers, and wikipedia refers to the entity set used.} More detailed results follow in the respective subsections, including precision, recall, and task-specific metrics.
\begin{table}[ht!]
    \centering
    \begin{adjustbox}{max width=\textwidth}
    \begin{tabular}{|c|c|c|c|} \hline
         \multicolumn{2}{|c|}{NER - UTFAA Strong F1}&  \multicolumn{2}{|c|}{CR - CoNLL-12 F1}\\ \hline
         \textbf{NLTK ne\_chunk}& \textbf{0.27}& \textbf{s2e-coref} & \textbf{0.8}\\ 
         flair (OntoNotes)&  0.22&  \textbf{ASP (flant5-large / t0-3b)} & \textbf{0.8}\\  
         spaCy EntityRecognizer (en\_core\_web\_lg)&  0.17&  neuralcoref (en\_core\_web\_lg)&  0.5\\
         stanza (ontonotes\_electra-large)&  0.16&  coref-mt5&  0.3 \\ \hline
         \multicolumn{2}{|c|}{NEL - Strong F1}&  \multicolumn{2}{|c|}{RE - Combined Acc} \\ \hline
         \textbf{spaCy EntityLinker (lg)}& \textbf{0.20}& \textbf{PL-Marker (albert-xxl)}&\textbf{1.0}\\
         BLINK (biencoder)&  0.14&  DeepStruct (NYT) & 0.7\\
         ReFinED (wiki\_w\_nums, wikipedia)&  0.10&  REBEL& 0.58 \\
         GENRE&  0.0063&  UniRel (NYT)& 0.083\\ \hline

    \end{tabular}
    \end{adjustbox}
    \caption{NLP Tool Zero-Shot Scores on FAA Data}
    \label{tab:Overview_Results}
\end{table}

\subsection{Named Entity Recognition}
As described in Section \ref{sec:NER_Eval}, we have strong-match and weak-match F1 scores for the NER tools which can be seen in the tables below. Table \ref{tab:NER_UTFAA} shows the label-agnostic results from evaluation on UTFAA, and Tables \ref{tab:NER_CoNLL}, \ref{tab:NER_ON}, \ref{tab:NER_ACE1}, and \ref{tab:NER_ACE05} show the SemEval F1 scores on the benchmark-annotated datasets. Each tool is evaluated on the benchmark-annotated dataset that corresponds to the set of entities it is trained to recognize, which is denoted in brackets. Note that we also provide scores for PL-Marker, since it performs NER as an intermediate step in RE, and outputs its named entities in a readily available file.

NLTK ne\_chunk is very sensitive to case. It can find entities well if given input with sentence-casing, including capitalized proper nouns. However, the FAA data is upper-cased by default, which loses the distinction between words that are naturally upper-cased or capitalized with those that are not. We input the data to ne\_chunk in lowercase and uppercase. It only recognized entities if the text was inputted in uppercase, and provided the label ``ORGANIZATION" for every one. This reduces trust in its overall effectiveness.

\begin{table}[h]
\centering
\begin{adjustbox}{max width=\textwidth}
\begin{tabular}{|l|c|c|c|c|c|c|}
\hline
& \multicolumn{3}{|c|}{\textbf{Weak}}& \multicolumn{3}{|c|}{\textbf{Strong}} \\
 & \textbf{Prec} & \textbf{Rec} & \textbf{F1} & \textbf{Prec} & \textbf{Rec} & \textbf{F1} \\ \hline
\textbf{PL-Marker (ACE-2005 bert)} & 0.78 & 0.27 & \textbf{0.4} & 0.68 & 0.24 & \textbf{0.35} \\ \hline
NLTK ne\_chunk (uppercased) & 0.43 & 0.37 & \textbf{0.4}& 0.29 & 0.25 & 0.27 \\ \hline
spaCy EntityRecognizer (en\_core\_web\_lg) & 0.6 & 0.16 & 0.26 & 0.39 & 0.11 & 0.17 \\ \hline
flair (OntoNotes) & 0.68 & 0.16 & 0.25 & 0.59 & 0.14 & 0.22 \\ \hline
PL-Marker (ACE-2005 albert-xxl) & 0.79 & 0.14 & 0.24 & 0.7 & 0.12 & 0.21 \\ \hline
spaCy EntityRecognizer (en\_core\_web\_sm) & 0.56 & 0.13 & 0.21 & 0.36 & 0.084 & 0.14 \\ \hline
stanza (ontonotes\_electra-large) & 0.73 & 0.1 & 0.18 & 0.66 & 0.094 & 0.16 \\ \hline
stanza (ontonotes-ww-multi\_electra-large) & 0.59 & 0.096 & 0.17 & 0.51 & 0.082 & 0.14 \\ \hline
stanza (ontonotes\_nocharlm) & 0.63 & 0.075 & 0.13 & 0.44 & 0.053 & 0.094 \\ \hline
PL-Marker (SciERC scibert-uncased) & 0.69 & 0.074 & 0.13 & 0.46 & 0.049 & 0.089 \\ \hline
flair (CoNLL-2003) & 0.77 & 0.064 & 0.12 & 0.64 & 0.053 & 0.098 \\ \hline
stanza (ontonotes-ww-multi\_nocharlm) & 0.49 & 0.071 & 0.12 & 0.38 & 0.055 & 0.096 \\ \hline
stanza (ontonotes\_charlm) & 0.71 & 0.066 & 0.12 & 0.51 & 0.047 & 0.086 \\ \hline
stanza (ontonotes-ww-multi\_charlm) & 0.64 & 0.045 & 0.084 & 0.42 & 0.029 & 0.055 \\ \hline
stanza (conll03\_charlm) & 0.54 & 0.04 & 0.075 & 0.42 & 0.031 & 0.059 \\ \hline
NLTK ne\_chunk (lowercased) & 0.0 & 0.0 & -- & 0.0 & 0.0 & -- \\ \hline
\end{tabular}
\end{adjustbox}
\caption{NER UTFAA Evaluation Results}
\label{tab:NER_UTFAA}
\end{table}

\begin{table}[ht!]
    \centering
    \begin{adjustbox}{max width=\textwidth}
    \begin{tabular}{|c|c|c|c|c|c|c|c|c|c|c|c|c|} \hline 
         & \multicolumn{3}{|c|}{\textbf{STRICT}}&\multicolumn{3}{|c|}{\textbf{EXACT}}&\multicolumn{3}{|c|}{\textbf{PARTIAL}}&\multicolumn{3}{|c|}{\textbf{TYPE}} \\
         & \textbf{Prec} & \textbf{Rec} & \textbf{F1} & \textbf{Prec} & \textbf{Rec} & \textbf{F1} & \textbf{Prec} & \textbf{Rec} & \textbf{F1} & \textbf{Prec} & \textbf{Rec} & \textbf{F1} \\ \hline
         \textbf{flair (CoNLL-03)} & 0.43 & 0.41 & \textbf{0.42} & 0.45 & 0.43 & \textbf{0.44} & 0.54 & 0.51 & \textbf{0.52} & 0.52 & 0.50 & \textbf{0.51} \\ \hline
         stanza (conll03\_charlm) & 0.21 & 0.18 & 0.20 & 0.24 & 0.20 & 0.22 & 0.33 & 0.28 & 0.30 & 0.34 & 0.30 & 0.32 \\ \hline
         \end{tabular}
    \end{adjustbox}
    \caption{NER CoNLLFAA Evaluation Results}
    \label{tab:NER_CoNLL}
\end{table}

\begin{table}[ht!]
    \centering
    \begin{adjustbox}{max width=\textwidth}
    \begin{tabular}{|c|c|c|c|c|c|c|c|c|c|c|c|c|} \hline 
         & \multicolumn{3}{|c|}{\textbf{STRICT}}&\multicolumn{3}{|c|}{\textbf{EXACT}}&\multicolumn{3}{|c|}{\textbf{PARTIAL}}&\multicolumn{3}{|c|}{\textbf{TYPE}} \\
         & \textbf{Prec} & \textbf{Rec} & \textbf{F1} & \textbf{Prec} & \textbf{Rec} & \textbf{F1} & \textbf{Prec} & \textbf{Rec} & \textbf{F1} & \textbf{Prec} & \textbf{Rec} & \textbf{F1} \\ \hline
         \textbf{stanza (ontonotes\_electra-large)} & 0.59 & 0.70 & \textbf{0.64} & 0.60 & 0.72 & \textbf{0.66} & 0.65 & 0.78 & \textbf{0.71} & 0.66 & 0.79 & \textbf{0.72} \\ \hline
         stanza (ontonotes-ww-multi\_electra-large) & 0.41 & 0.55 & 0.47 & 0.42 & 0.56 & 0.48 & 0.49 & 0.66 & 0.57 & 0.51 & 0.68 & 0.58 \\ \hline
         flair (OntoNotes) & 0.32 & 0.61 & 0.42 & 0.36 & 0.68 & 0.47 & 0.40 & 0.77 & 0.53 & 0.40 & 0.76 & 0.52 \\ \hline
         stanza (ontonotes\_charlm) & 0.32 & 0.25 & 0.28 & 0.34 & 0.26 & 0.30 & 0.44 & 0.34 & 0.38 & 0.43 & 0.33 & 0.37 \\ \hline
         stanza (ontonotes\_nocharlm) & 0.23 & 0.22 & 0.23 & 0.31 & 0.30 & 0.31 & 0.41 & 0.40 & 0.40 & 0.30 & 0.29 & 0.29 \\ \hline
         stanza (ontonotes-ww-multi\_nocharlm) & 0.22 & 0.25 & 0.23 & 0.26 & 0.30 & 0.28 & 0.36 & 0.42 & 0.39 & 0.28 & 0.33 & 0.31 \\ \hline
         stanza (ontonotes-ww-multi\_charlm) & 0.31 & 0.18 & 0.22 & 0.36 & 0.21 & 0.27 & 0.57 & 0.33 & 0.42 & 0.56 & 0.32 & 0.41 \\ \hline
         spaCy EntityRecognizer (en\_core\_web\_sm) & 0.10 & 0.19 & 0.13 & 0.18 & 0.35 & 0.24 & 0.24 & 0.46 & 0.31 & 0.16 & 0.31 & 0.21 \\ \hline
         spaCy EntityRecognizer (en\_core\_web\_lg) & 0.071 & 0.16 & 0.098 & 0.14 & 0.32 & 0.20 & 0.18 & 0.41 & 0.25 & 0.092 & 0.21 & 0.13 \\ \hline
        \end{tabular}
    \end{adjustbox}
    \caption{NER ONFAA Evaluation Results}
    \label{tab:NER_ON}
\end{table}

\begin{table}[ht!]
    \centering
    \begin{adjustbox}{max width=\textwidth}
    \begin{tabular}{|c|c|c|c|c|c|c|c|c|c|c|c|c|} \hline 
         & \multicolumn{3}{|c|}{\textbf{STRICT}}&\multicolumn{3}{|c|}{\textbf{EXACT}}&\multicolumn{3}{|c|}{\textbf{PARTIAL}}&\multicolumn{3}{|c|}{\textbf{TYPE}} \\
         & \textbf{Prec} & \textbf{Rec} & \textbf{F1} & \textbf{Prec} & \textbf{Rec} & \textbf{F1} & \textbf{Prec} & \textbf{Rec} & \textbf{F1} & \textbf{Prec} & \textbf{Rec} & \textbf{F1} \\ \hline
         NLTK ne\_chunk (uppercased) & 0.0066 & 0.022 & 0.01 & 0.087 & 0.30 & 0.13 & 0.13 & 0.44 & 0.20 & 0.024 & 0.081 & 0.037 \\ \hline
         NLTK ne\_chunk (lowercased) & 0.00 & 0.00 & -- & 0.00 & 0.00 & -- & 0.00 & 0.00 & -- & 0.00 & 0.00 & -- \\ \hline
        \end{tabular}
    \end{adjustbox}
    \caption{NER ACE1FAA Evaluation Results}
    \label{tab:NER_ACE1}
\end{table}

\begin{table}[ht!]
    \centering
    \begin{adjustbox}{max width=\textwidth}
    \begin{tabular}{|c|c|c|c|c|c|c|c|c|c|c|c|c|} \hline 
         & \multicolumn{3}{|c|}{\textbf{STRICT}}&\multicolumn{3}{|c|}{\textbf{EXACT}}&\multicolumn{3}{|c|}{\textbf{PARTIAL}}&\multicolumn{3}{|c|}{\textbf{TYPE}} \\
         & \textbf{Prec} & \textbf{Rec} & \textbf{F1} & \textbf{Prec} & \textbf{Rec} & \textbf{F1} & \textbf{Prec} & \textbf{Rec} & \textbf{F1} & \textbf{Prec} & \textbf{Rec} & \textbf{F1} \\ \hline
         \textbf{PL-Marker (ACE-2005 bert)} & 0.53 & 0.47 & \textbf{0.50} & 0.54 & 0.49 & \textbf{0.51} & 0.67 & 0.60 & \textbf{0.64} & 0.77 & 0.69 & \textbf{0.73} \\ \hline
         PL-Marker (ACE-2005 albert-xxl) & 0.62 & 0.28 & 0.39 & 0.62 & 0.28 & 0.39 & 0.78 & 0.35 & 0.49 & 0.92 & 0.42 & 0.58 \\ \hline
    \end{tabular}
    \end{adjustbox}
    \caption{NER ACE05FAA Evaluation Results}
    \label{tab:NER_ACE05}
\end{table}

\subsection{Coreference Resolution}
As described in Section \ref{sec:CR_Eval}, we have four different F1 scores for CR. Additionally, we have the CoNLL-2012 F1, which is the unweighted average of the F1 scores from MUC, B-CUBED, and CEAF. See Table \ref{tab:CR_Results} for results. ASP flant5-large and t0-3b happen to have the same output on our evaluation sample; their repeated scores are not a mistake. We only report one significant figure because there are only eighteen records with any coreferences in our gold standard.

\begin{table}[ht!]
    \centering
    \begin{adjustbox}{max width=\textwidth}
    \begin{tabular}{|c|c|c|c|c|c|c|c|c|c|c|c|c|c|} \hline
         & \multicolumn{3}{c|}{\textbf{MUC}}& \multicolumn{3}{c|}{\textbf{B-cubed}}& \multicolumn{3}{c|}{\textbf{CEAF}}&\textbf{CoNLL-12}& \multicolumn{3}{c|}{\textbf{LEA}} \\
         & \textbf{Prec} & \textbf{Rec} & \textbf{F1} & \textbf{Prec} & \textbf{Rec} & \textbf{F1} & \textbf{Prec} & \textbf{Rec} & \textbf{F1} & \textbf{F1} & \textbf{rec} & \textbf{Rec} & \textbf{F1} \\ \hline
         \textbf{s2e-coref} & 0.9& 0.7& \textbf{0.8}& 0.9& 0.7& \textbf{0.8}& 0.9& 0.7& \textbf{0.8}& \textbf{0.8}& 0.9& 0.7& \textbf{0.8}\\ \hline
         \textbf{ASP (t0-3b)}& 0.7& 0.7& 0.7& 0.7& 0.8& \textbf{0.8}& 0.8& 0.8& \textbf{0.8}& \textbf{0.8}& 0.7& 0.7&0.7\\ \hline
         \textbf{ASP (flant5-large)} & 0.7& 0.7& 0.7& 0.7& 0.8& \textbf{0.8}& 0.8& 0.8& \textbf{0.8}& \textbf{0.8}& 0.7& 0.7& 0.7\\ \hline
         ASP (flant5-xl) & 0.7& 0.6& 0.7& 0.8& 0.6& 0.7& 0.8& 0.6& 0.7& 0.7& 0.7& 0.6& 0.6\\ \hline
         ASP (flant5-base) & 0.7& 0.5& 0.6& 0.7& 0.5& 0.6& 0.7& 0.6& 0.6& 0.6& 0.7& 0.5& 0.6\\ \hline
         neuralcoref (en\_core\_web\_lg) & 0.6& 0.4& 0.5& 0.6& 0.4& 0.5& 0.6& 0.5& 0.5& 0.5& 0.6& 0.4& 0.5\\ \hline
         coref\_mt5 & 0.8& 0.2& 0.3& 0.9& 0.2& 0.3& 0.8& 0.2& 0.3& 0.3& 0.8& 0.2& 0.3\\ \hline
         neuralcoref (en\_core\_web\_sm)& 0.1& 0.1& 0.1& 0.2& 0.1& 0.1& 0.3& 0.2& 0.2& 0.1& 0.1& 0.1& 0.1\\ \hline
    \end{tabular}
    \end{adjustbox}
    \caption{Coreference Resolution Quantitative Evaluation Results}
    \label{tab:CR_Results}
\end{table}

\subsection{Named Entity Linking}
As described in Section \ref{sec:NEL_Eval}, we have three different evaluations for NEL, strong-matching, weak-matching, and flexible. We show the results for each evaluation below.\footnote{For ReFinED, wiki\_w\_nums is an abbreviation for the model, wikipedia\_model\_with\_numbers, wiki is an abbreviation for the model wikipedia\_model, and aida is an abbreviation for the model aida\_model. Wikipedia and Wikidata refer to the entity sets used in inference.} Tables \ref{tab:NEL_F1_Results} compares tools' performance on an F1 metric, and Table \ref{tab:NEL_OT_Results} compares tools' performance on the OT metrics, JC and Class similarity. Note that the F1 metric results in a very different ranking than the OT metrics. This is due to the influence of recall on F1. spaCy EntityLinker has a high recall on the entities in our GS, but does not correctly link them at as high a rate than most of the other tools. Since the OT metrics only compare valid predicted entities against the gold standard, missing gold standard entities have no bearing on the score.

Note that the results for ReFinED's aida\_model are close to zero for F1 scores since it only recognizes and links eight entities in the sample records. The OT scores, then, should be understood as circumstantial to the eight entities found, and not a reliable indication of the aida\_model's performance.

\begin{table}[ht!]
    \centering
    \begin{adjustbox}{max width=\textwidth}
    \begin{tabular}{|c|c|c|c|c|c|c|c|c|c|} \hline
         & \multicolumn{3}{|c|}{\textbf{Weak}}& \multicolumn{3}{|c|}{\textbf{Strong}} & \multicolumn{3}{|c|}{\textbf{Flex}}\\
         & \textbf{Prec} & \textbf{Rec} & \textbf{F1} & \textbf{Prec} & \textbf{Rec} & \textbf{F1} & \textbf{Prec} & \textbf{Rec} & \textbf{F1}\\ \hline
         \textbf{spaCy EntityLinker (en\_core\_web\_lg)} & 0.19 & 0.20 & \textbf{0.19}& 0.15 & 0.30 & \textbf{0.20}& 0.15& 0.18& \textbf{0.16} \\ \hline
         spaCy EntityLinker (en\_core\_web\_sm) & 0.17 & 0.21 & \textbf{0.19}& 0.14 & 0.30 & \textbf{0.19}& 0.15 & 0.17 & \textbf{0.16}\\ \hline
         BLINK (biencoder) & 0.64 & 0.068 & 0.12 & 0.57 & 0.08 & 0.14 & 0.64& 0.052& 0.096 \\ \hline
         BLINK (crossencoder) & 0.61 & 0.065 & 0.12 & 0.52 & 0.074 & 0.13 & 0.61& 0.05 & 0.092 \\ \hline
         ReFinED (wiki\_w\_nums, wikipedia)& 0.67 & 0.058 & 0.11 & 0.53 & 0.056 & 0.10 & 0.71& 0.049& 0.092\\ \hline
 ReFinED (wiki\_w\_nums, wikidata)& 0.72 & 0.058 & 0.11 & 0.57 & 0.056 & 0.10 & 0.76& 0.049&0.092\\ \hline
         ReFinED (wiki, wikidata)& 0.35& 0.03& 0.055& 0.30& 0.041& 0.073& 0.36& 0.028& 0.051\\ \hline
         ReFinED (wiki, wikipedia)& 0.33 & 0.03 & 0.055 & 0.29 & 0.041 & 0.073 & 0.35& 0.028& 0.051\\ \hline
         GENRE & 0.12 & 0.0032 & 0.0063 & 0.059 & 0.0033 & 0.0063 & 0.20 & 0.0045 & 0.0087\\ \hline
 ReFinED (aida, wikidata)& 0.0& 0.0& 0.0& 0.17& 0.0032& 0.0063& 0.0& 0.0&0.0\\ \hline
 ReFinED (aida, wikipedia)& 0.0& 0.0& 0.0& 0.0& 0.0& 0.0& 0.0& 0.0&0.0\\ \hline
    \end{tabular}
    \end{adjustbox}
    \caption{NEL F1 Quantitative Evaluation Scores}
    \label{tab:NEL_F1_Results}
\end{table}

\begin{table}[ht!]
    \centering
    \begin{adjustbox}{max width=\textwidth}
    \begin{tabular}{|c|c|c|c|c|c|c|} \hline
         & \multicolumn{2}{|c|}{\textbf{Weak}}& \multicolumn{2}{|c|}{\textbf{Strong}} & \multicolumn{2}{|c|}{\textbf{Flex}}\\
         & JC& Class& JC & Class& JC & Class\\ \hline
         \textbf{ReFinED (wiki\_w\_nums, wikipedia)} & \textbf{0.89}& \textbf{0.87} & \textbf{0.84}& \textbf{0.81} & \textbf{0.90}& \textbf{0.88}\\ \hline
         \textbf{ReFinED (wiki\_w\_nums, wikidata)} & \textbf{0.89}& \textbf{0.87} & 0.80& 0.78& \textbf{0.90}& \textbf{0.89}\\ \hline
         ReFinED (wiki, wikidata)& 0.84& 0.74& 0.74& 0.65& 0.82& 0.72\\ \hline
 ReFinED (aida, wikidata)& 0.84& 0.48& 0.73& 0.5& 0.61&0.33\\ \hline
 BLINK (biencoder) & 0.82 & 0.77 & 0.68 & 0.64  & 0.8&0.75\\ \hline
 BLINK (crossencoder) & 0.82 & 0.77 & 0.64 & 0.60 & 0.8&0.75\\\hline
         ReFinED (wiki, wikipedia)& 0.80& 0.71& 0.71& 0.62& 0.78& 0.69\\ \hline
         ReFinED (aida, wikipedia)& 0.56& 0.32& 0.49& 0.25& 0.46& 0.25\\ \hline
         GENRE & 0.32 & 0.28 & 0.27 & 0.20 & 0.37 & 0.33 \\ \hline
         spaCy EntityLinker (en\_core\_web\_sm) & 0.15 & 0.073 & 0.14 & 0.071 & 0.15 & 0.082\\ \hline
         spaCy EntityLinker (en\_core\_web\_lg) & 0.15 & 0.082 & 0.13 & 0.069 & 0.14 & 0.077 \\ \hline
    \end{tabular}
    \end{adjustbox}
    \caption{NEL OT Quantitative Evaluation Scores}
    \label{tab:NEL_OT_Results}
\end{table}

\subsection{Relation Extraction}
As described in Section \ref{sec:RE_Eval},
and \ref{sec:RE_Eval_Guidelines}, 
we qualitatively evaluated RE for syntactic accuracy, semantic accuracy, and consistency. These scores can be found in Table \ref{tab:RE_Results}, which orders the tools by number of triples evaluated, and highlights high-scorers in bold. Note that the rightmost column ``\% Docs w/ Predicted Trip" denotes the percentage of the 2748 OMIn dataset records for which the corresponding tool extracted one or more triples.

Since REBEL extracts at least ten times more triples than the other tools, its syntactic, semantic, and consistency numbers are more meaningful. This in part due to REBEL's set of 220 relations, which is far larger than the sets of 6-25 relations on which the other RE tools are trained. However, the 6 most common relations make up 60\% of the generated triples and the 25 most common relations make up 92\% of the generated triples, so the greater number of possible relations does not fully account for the high output. Additionally, we found that REBEL generated an entity with no matching textual mention three times in the sample, suggesting that it also hallucinates on rare occasions.

Although PL-Marker had very low output, it generated notably reliable and sensible results. Its success is qualified by the fact that the relations it generates, which are from SciERC and ACE-2005, are much more broadly defined and have fewer syntactic rules than those in REBEL or UniRel. If used in a KE workflow, the resultant KG would be much less precise and informative than one created with more strongly defined relations, such as the Wikidata properties used in REBEL.

Additionally, we report a combined accuracy, which is an unweighted average of syntactic and semantic accuracy, as in (\cite{yang2021measuring}).

\begin{table}[ht!]
    \centering
    \begin{adjustbox}{max width=\textwidth}
    \begin{tabular}{|c|c|c|c|c|c|c|c|} \hline
         &  \textbf{\# Trip Eval'd}&  \textbf{Syn Acc}&  \textbf{Sem Acc}&  \textbf{Con}& \textbf{Combined Acc}&  \textbf{\# Trip's}& \textbf{\% Docs w/ Predicted Trip}\\ \hline 
         REBEL&  181&  0.86&  0.30&  0.99&0.58&  4766& 99\\ \hline 
         PL-Marker (bert) RE&  18&  \textbf{1.0}&  0.94&  1.0&0.97&  289& 10\\ \hline 
         PL-Marker (scibert) RE&  9&  \textbf{1.0}&  0.56&  1.0 &0.78&  127& 4\\ \hline 
 PL-Marker (albert-xxl) RE& 4& \textbf{1.0}& \textbf{1.0}& 1.0 & \textbf{1.0}& 147&4\\\hline
         Unirel (NYT)&  6&  0.17&  0.0&  1.0 &0.083&  87& 2\\ \hline 
         DeepStruct (NYT)&  5&  0.80&  0.60&  1.0&0.7&  93& 3\\ \hline
    \end{tabular}
    \end{adjustbox}
    \caption{Relation Extraction Qualitative Evaluation Results}
    \label{tab:RE_Results}
\end{table}

Since UniRel and PL-Marker's scibert and albert-based models return so few triples on the 100 sample records, we also perform a supplementary evaluation of all the triples predicted over the complete OMIn dataset. These scores can be seen in Table \ref{tab:RE_All_Results}.

\begin{table}[ht!]
\centering
\begin{adjustbox}{max width=\textwidth}
\begin{tabular}{|c|c|c|c|c|c|}\hline
&  \textbf{\# Trip's Eval'd}&  \textbf{Syn Acc}&  \textbf{Sem Acc}&  \textbf{Con}& \textbf{Combined Acc}\\ \hline 
PL-Marker (albert-xxl) RE & 158 & \textbf{0.98}& \textbf{0.98}& 1.0 & \textbf{0.98}\\ \hline 
PL-Marker (scibert) RE & 127 & \textbf{1.0}& 0.78 & 1.0 & 0.89\\ \hline
 DeepStruct (NYT)& 93& 0.79& 0.65& 1.0 & 0.72\\\hline
UniRel (NYT) & 87 & 0.37& 0.14& 0.98 & 0.26\\ \hline
\end{tabular}
\end{adjustbox}
\caption{Supplementary Relation Extraction Evaluation}
\label{tab:RE_All_Results}
\end{table}

Lastly, since the PL-Marker's bert-based model returned very few triples in the evaluation sample but too many to evaluate by hand, we selected at random 1000 records from the OMIn dataset and qualitatively evaluated the resulting triples. These scores are available in Table \ref{tab:RE_1000_Results}.

\begin{table}[ht!]
\centering
\begin{adjustbox}{max width=\textwidth}
\begin{tabular}{|c|c|c|c|c|c|}\hline
 &  \textbf{\# Trip's Eval'd}&  \textbf{Syn Acc}&  \textbf{Sem Acc}&  \textbf{Con}& \textbf{Combined Acc}\\ \hline 
PL-Marker (bert) RE & 126 & 0.96 & 0.99 & 1.0 & 0.98 \\ \hline 
\end{tabular}
\end{adjustbox}
\caption{Supplementary Relation Extraction Evaluation, PL-Marker BERT Model}
\label{tab:RE_1000_Results}
\end{table}

\section{Discussion}

\subsection{Performance}
The selected tools scored significantly lower on the OMIn dataset than on benchmark datasets (in the majority of cases). Some notable exceptions are the CR tools, s2e-coref and ASP, and the RE tools, PL-Marker and REBEL. Both NER and NEL tools failed to reliably extract GS entities. NER tools, in general, performed much better on benchmark-annotated GSs, but still significantly below reported scores for those benchmarks, which indicates that they struggle to transfer to the maintenance domain. Additionally, we found that errors in identifying entity spans often arose in sentences with uncommon syntax and shorthand, acronyms and abbreviations were often ignored or misidentified, and the overall efficacy of knowledge extraction was limited by the prevalence of omitted subjects in sentences.

\subsection{Trust}
In this work, we focused on trust in four facets:
\begin{itemize}
    \item{\textbf{Privacy and Confidentiality}} None of the NLP nor LLM models we evaluated were allowed to leverage data storage or APIs external to our private testing infrastructure. This allows an organization to keep their confidential information private.
    \item{\textbf{Accuracy and Robustness}} Each tool's knowledge extraction capability was evaluated to determine what level of accuracy and organization could expect from an NLP tool or LLM not trained or tuned for their domain.  Evaluating tools in diverse domains is important to understand robustness.
\item{\textbf{Reproducibility}} It is essential that results from the tools selected can be reproduced and do not vary from test to test. Since we repeatedly evaluated zero-shot scores for 16 different tools, we ensured that results were not influenced by previous data passed into the tools and models.  Further, we evaluated the degree of complexity to build and run the tools as discussed more in the following section.
\item{\textbf{Accountability}} We selected and our gold standard dataset and evaluation metrics based on peer reviewed community standards.  We then documented all of the processes and procedures either directly in this work, its appendix or the public OMIn data repository.
\end{itemize}

\subsection{Technology Readiness Level}
The Technology Readiness Levels (TRLs) provide a 9-level gauge of how close a tool is to being ready for launch. The scale ranges from basic technology research at level 1 to system proven in an operational environment at level 9 (\cite{mankins1995technology}). Because TRLs communicate technological maturity in the context of a target operational environment, TRL assessments provide a way for stakeholders in operational environments to shape future research and development.
We measured the TRLs of the sixteen tools based on their performance on FAA maintenance data. Because the F1 and accuracy scores are very low, the TRL levels are in the 1-2 range (basic technology research and research to produce feasibility) as shown in Table \ref{tab:TRL_Levels}.
Some tools required preprocessing and did not let FAA data to be passed in directly as text. Additionally, some tools were either outdated, had complex software version dependencies, or did not have clear documentation explaining how to run the tools. This in turn lowered the TRL level rating for the respective tool. A ``Reproducibility Rating" for each tool can be found in our ReadMEs on GitHub, which describes the challenges we encountered during the setup process.

\begin{table}[ht!]
    \centering
    \begin{adjustbox}{max width=\textwidth}
    \begin{tabular}{|c|c|c|c|c|c|c|c|} \hline
         \multicolumn{2}{|c|}{NER}&  \multicolumn{2}{|c|}{Coref}&  \multicolumn{2}{|c|}{NEL}&  \multicolumn{2}{|c|}{RE}\\ \hline 
         spaCy&  2&  ASP&  1&  BLINK& 2&  REBEL& 1\\ \hline 
         flair&  1&  coref-mt5&  1&  spaCy Entity Linker&  2&  UniRel& 1\\ \hline 
         stanza&  2&  s2e-coref& 1&  GENRE&  1&  DeepStruct& 1\\ \hline 
         nltk& 1&  neuralcoref& 1&  ReFinED&  2&  PL-Marker& 2\\ \hline
    \end{tabular}
    \end{adjustbox}
    \caption{Tool TRL Levels for Zero-Shot on FAA Data}
    \label{tab:TRL_Levels}
\end{table}

\section{Conclusion}

We introduced the Operations and Maintenance Intelligence (OMIn) dataset that provides a curated subset of FAA Accident/Incident narratives with gold standards for NER, CR, and NEL to support knowledge extraction in maintenance settings. OMIn pairs terse, shorthand-heavy text with structured fields (e.g., aircraft details, failure codes, dates) and, to our knowledge, constitutes the first open-source KE resource focused on operations and maintenance. Using OMIn, we conducted a zero-shot, on-premises evaluation of sixteen openly available tools across NER, CR, NEL, and RE. The results show that off-the-shelf systems struggle with the abbreviations, non-standard syntax, and short narratives typical of maintenance text: NER and NEL exhibit notably low recall, CR transfers more robustly, and RE methods display a precision–coverage trade-off. Together with our reproducibility observations, these findings suggest low technology readiness for immediate operational deployment without domain adaptation. OMIn v1 is derived from pilot/crew narratives rather than technician logs, and we do not provide a unified RE gold standard due to heterogeneous relation schemas; these are deliberate scope choices for a first baseline.

These outcomes directly motivate future work: normalization and acronym expansion, enriched domain-specific labels, and larger gold standards are the most immediate levers to raise recall and comparability. A shared maintenance ontology would enable a fair RE benchmark and support joint models that fuse text with OMIn’s structured fields. Resources will be released via our GitHub repository and an anonymized Zenodo DOI (withheld for review), and we invite community contributions to iterate OMIn and its benchmarks.

% Stop adding to the TOC
\addtocontents{toc}{\protect\setcounter{tocdepth}{-1}}

\section{Declaration of generative AI and AI-assisted technologies in the writing process}
During the preparation of this work, the authors used the web
interfaces of GPT-4o, https://chatgpt.com/, in order to improve the readability and flow of this manuscript. After using this service, the authors reviewed and edited the content as needed and take full responsibility for the content of the publication.

\section{Acknowledgments}
We would like to thank University of Notre Dame students Danny Finch, Alyssa Riter, and Lindsey Michie for their contributions to curating the OMIn baseline data set. We would also like to thank Crane NSWC Technical Points of Contact (TPOCs) Alicia Scott, Eli Phillips, Aimee Flynn, Adam Shull, and Tim Kelley for their insight into operational priorities and challenges related to trusted MO. Professor Christopher Sweet provided consultation on metric selection and development. The Notre Dame Center for Research Computing and the SCALE Program at Purdue University provided funding support.

%% If you have bib database file and want bibtex to generate the
%% bibitems, please use
\Urlmuskip=0mu plus 1mu\relax
\bibliographystyle{elsarticle-harv} 
\bibliography{ie-paper.bib}
%% Refer following link for more details about bibliography and citations.
%% https://en.wikibooks.org/wiki/LaTeX/Bibliography_Management

\newpage

\appendix

\section{CoNLL-2012 Format Pre-Processing}
\label{sec:CoNLL-2012_Pre-Processing}

For CoNLL-2012, documents were organized in a tabular structure with each word on a separate line and more than 11 columns detailing the word's semantic role. These columns include parts of speech, parse bits, predicate lemmas, speakers, and named entities. The parse bit links each word to a segment of the sentence's Propbank-style parse tree, originally generated using a Charniak parser by the CoNLL-2012 developers. To recreate their process, we used the Charniak parser from Brown Laboratory for Linguistic Information Processing (BLLIP), along with the NLTK POS-tagger and spaCy EntityRecognizer. The CoNLL-2012 developers used the Identifinder\textsuperscript{TM} tool from BBN (\cite{miller_bbn_1998}) for NER; however, we used spaCy since it was more up-to-date. All records were uniformly labeled with `speaker1' in the speaker field. We found through experimentation that the selected CoNLL-2012-based tools did not use the predicate-lemma and its associated columns, so we inserted placeholders in these fields. The OMIn data formatted for CoNLL-2012 is available in our repository (\cite{Mealey_Operations_and_Maintenance_2024}).

Although we aimed to replicate the original CoNLL-2012 dataset from OntoNotes 5.0, discrepancies in the POS-tagging, NER, and parsing tools led to a less-than-perfect match. Without access to the original tools, our version serves as a close approximation.

\section{NER Annotation Guidelines}
\label{sec:NER_Annotation_Guidelines}

We follow ACE-2005 to label persons, locations, organizations, geo-political entities (GPEs), facilities, weapons, and vehicles. We make a few exceptions. First, we include "ground" and "land" as location entities, since they are distinct locations in aviation, where they are often used to differentiate from airspace. We also exclude articles from our entities, but keep all other modifiers. Lastly, we do not include relative clauses or relative pronouns in our GS, since they are unhelpful as a basis for NEL.

We follow OntoNotes 5.0 to label dates, times, quantities, and cardinals.

Additionally, we label entities that fall into one of the following categories: vehicle system/component, operational items (fuel, oil, load, etc.), failures, causes of failures, symptoms of failures, phases of flight (takeoff, climb, landing, etc.), types of flight (ferry flight, test flight, etc.), and procedures (maintenance, safety checks, etc.). Future work could involve formalizing these categories into well-defined entity types. We follow ACE-2005 in all syntactical rules such as the inclusion of modifying phrases (except for articles), nesting entities, treating appositives, etc.

We ignore all typos and words which are cut off at the end of the record. However, we include shorthand and acronyms (``acft" for aircraft, ``prop" for propeller, etc.).

\paragraph{Example 1: Nested Entities}

For the record, ``(-23) Mr. Timothy Allen Wells was acting as pilot in command of a Bell Helicopter model BHT-47-G5, N4754R, engaged", we follow ACE-2005's nested entity guidelines and include ``Mr. Timothy Allen Wells", ``pilot in command of a Bell Helicopter model BHT-47-G5, N4754R", ``Bell Helicopter model BHT-47-G5, N4754R", ``Bell Helicopter model BHT-47-G5", ``Bell", and ``N4754R".

\paragraph{Example 2: Aviation Entities}

For the record, ``After departing high oil temp. Landed off airpor. Sheared main gear. Found low on oil.", the entities are ``high oil temp",``oil",``sheared main gear", and ``OIL". The first ``OIL" is an entity nested in ``high oil temp". ``oil" is included because it is an operational item, ``high oil temp" is included because it is a failure or a failure symptom, and ``sheared main gear"" is both a failure and a system component. Since we do not label the entities with definite types, the ambiguity of ``sheared main gear"'s type is not an issue.

\section{NEL Evaluation Guidelines}
\label{sec:NEL_Eval_Appendix}

\subsection{F1 Score Details}
We define a true positive as a predicted entity that matches both the entity and the QID in a gold standard link. A false positive is a predicted entity that matches an entity but not its QID in a gold standard link. A false negative exists when there is no matching predicted entity for a gold standard entity-QID link. Predicted entities without any QID, as well as predicted entities-QID links without a matching gold entity, are not included in the evaluation.

\subsection{OT Metrics Details}
For the OT metrics, we evaluate the intersection of entities in the gold and predicted sets. This intersection consists of predicted entities that have a matching entity in the gold standard, where both predicted and gold entities are linked with a QID. We then report a micro-average across all linked entities in the intersection, excluding those for which a score could not be obtained due to limitations in the KB.

\textit{Jiang Conrath }(JC) is an information-theoretic distance metric that combines path-based features with information content to provide a nuanced similarity metric~\cite{jiang1997semanticsimilaritybasedcorpus}. The formula is given by:
\begin{equation} 
\text{jc}(c_1, c_2) = 2 \cdot \log p(\text{mss}(c_1, c_2)) - (\log p(c_1) + \log p(c_2)) 
\end{equation}
where \( \text{jc}(c_1, c_2) \) denotes the distance between concepts $c_1$ and $c_2$, \( \text{mss}(c_1, c_2) \) is the most specific subsumer of $c_1$ and $c_2$, and $p(c)$ is the probability of encountering an instance of concept $c$. 
In \cite{kgtk_similarity}, they use instance counts of a class to compute the probability $p(c)$ and normalize Jiang Conrath distance onto a $[0\ldots1]$ similarity measure by dividing by the largest possible distance between $c_1$ and $c_2$ through the \textit{entity} node (Q351201) in the ontology.

The \textit{class} similarity computation employs the Jaccard Similarity of the superclass sets of two nodes, inversely weighted by the instance counts of the classes. Formally, for two concepts \(c_1\) and \(c_2\), let \(S(c_1)\) and \(S(c_2)\) represent their respective sets of superclasses, and let \(I(c)\) denote the instance count of class \(c\). The class similarity is defined as:

\begin{equation}
\text{class\_sim}(c_1, c_2) = \frac{\sum_{c \in S(c_1) \cap S(c_2)} \frac{1}{I(c)}}{\sum_{c \in S(c_1) \cup S(c_2)} \frac{1}{I(c)}}
\end{equation}
where the term \( \frac{1}{I(c)} \) inversely weights each class by its instance count, thereby reducing the influence of more general classes with higher instance counts and emphasizing more specific classes.

Examples of JC and class similarity computations are presented in Table \ref{tab:OT_samples}, which shows the similarity values between the concept $c_1$ = \textit{aviation fuel} (Q1875633) and other $c_2$ concepts, such as \textit{aviation fuel} (Q1875633), \textit{combustible matter} (Q42501), \textit{Fuel} (Q15766923), and \textit{Fuel} (Q5507117). Note that although some concepts include the word \textit{Fuel} in their label, they receive the lowest scores due to their semantic distance from the $c_1$ concept, \textit{aviation fuel}, as indicated by their values under the column description.

\begin{table}[ht!]
\centering

\renewcommand{\arraystretch}{1.2}
\scriptsize
%\begin{adjustbox}{max width=\textwidth}
\begin{tabular}{|L{1.45cm}|L{6.1cm}|L{2.6cm}|C{.79cm}|C{.6cm}|}\hline
\textbf{QID} & \textbf{Description} & \textbf{Label} & \textbf{Class} & \textbf{JC} \\ \hline 
Q1875633 & propellents used to power aircraft or aviation...& aviation fuel & 1.00 & 1.00 \\ \hline 
Q42501 & any material that stores energy that can... & combustible matter & 0.68 & 0.89 \\ \hline 
Q15766923 & scientific journal & Fuel & 0.03 & 0.06 \\ \hline 
Q5507117 & short-lived Bay Area post-hardcore musical...& Fuel & 0.00 & 0.00 \\ \hline
\end{tabular}
%\end{adjustbox}

\caption{Example OT Measurements}
\label{tab:OT_samples}
\end{table}

\subsection{Flexible Evaluation}
Flexible evaluation makes use of the Flexible NEL GS, as described in Section \ref{sec:NEL_GS}. The Flexible GS includes secondary and tertiary linked entities as well as the primary linked entities used in the other evaluation strategies. In Flexible evaluation, if a predicted linked entity exactly matches either the primary, secondary, or tertiary link, it is correct. Flexible evaluation utilizes strong-matching.

Example: ``Forward cargo door opened as aircraft took off. Objects dropped out. Returned. Failed to see warning light."

The primary, secondary, and tertiary entities are laid out in Table \ref{tab:Flexible_GS_Ex}. In strong and weak-matching evaluation, only (``aircraft",Q11436) would be included in the gold standard. In flexible evaluation, (``door", Q36794), (``aircraft",Q11436), and (``light", Q1146001) would be included.

\begin{table}[ht!]
\centering
\begin{adjustbox}{max width=\textwidth}
\begin{tabular}{|c|c|c|c|c|c|}\hline
\textbf{Primary Ent}& \textbf{Primary QID}& \textbf{Secondary Ent}& \textbf{Secondary QID}& \textbf{Tertiary Ent}& \textbf{Tertiary QID}\\ \hline 
forward cargo door& & cargo door& & door& Q36794\\ \hline 
aircraft& Q11436& & & & \\ \hline 
warning light& & light& Q1146001& & \\ \hline

\end{tabular}
\end{adjustbox}
\caption{Example NEL Flexible Entities and QIDs}
\label{tab:Flexible_GS_Ex}
\end{table}

Secondary entities are also linked to primary QIDs when available, and so too with tertiary entities to secondary and primary QIDs. This is done so that if a tool links a more ``general" mention to the QID for the fitting, context-specific Wikidata entity, rather than the general QID, it is not penalized. For example, the GS for a document in the FAA data includes the primary link (``forced landing", Q1975745) and the secondary link (``landing", Q844947). If a tool predicted (``landing",Q1975745), that would be counted as correct, since it inferred from context that it was a forced landing and linked it to the corresponding QID.

\section{RE Evaluation Guidelines}
\label{sec:RE_Eval_Guidelines}

\subsection{Syntactic Accuracy}
Syntactic accuracy is the degree to which a tool's output follows the grammatical rules in our set of guidelines. A triple is either completely syntactically accurate (1.0), half syntactically accurate (0.5), or syntactically inaccurate (0.0), depending on whether both, one of, or neither of the head and tail entities are correct, respectively. Our guidelines are recorded below:

\begin{itemize}
\item Head and tail entities must consist of complete phrases. ``Complete phrase" signifies a word or phrase which can be treated as a noun or a verb. For example, the triple (``cowling",``part of",``engine in") is inaccurate, since ``engine in" is not a complete phrase.
\item If a word or phrase is used as a modifier in a sentence (and is thus not its own phrase in that particular sentence), it may still be counted as a complete phrase if it can function as a noun, verb, noun phrase, or verb phrase in another context. For example, in the sentence ``Wing fuel tank sumps were not drained during preflight", (``sumps",``part of",``wing fuel tank") and (``fuel tank sumps", ``part of", ``wing") would both be syntactically accurate.
\item \textbf{Exception} to the above two rules: personal pronouns may be entities, and should be interpreted the same as the person they refer to.
\item If a head or tail entity includes words or phrases that modify a part of the sentence outside of that included in the entity, it is inaccurate. For example, in the sentence ``Engine cowling separated from engine in flight," the subject is ``engine cowling", and the verb is ``separated", modified by ``in flight" and ``from engine." Because ``in flight" modifies "separated," the predicted entity "engine in flight" would be syntactically inaccurate.
\item Verbs and verb phrases may only be used as entities if the relation can accept an event-type entity. Verb phrases also do not need to have a subject. For example: (``improper preflight", ``has effect", ``crashed") is syntactically accurate.
\item Complete clauses (subject-verb) may only be used as entities if the relation can accept an event-type entity.
\item A head entity may be a subspan of its tail entity, and vice versa.
\item Head and tail entities must follow syntax constraints implied by the relation. For example, the relation "place of birth" must have a location as the head and a person as the tail. These constraints are described for each set of relations below under Syntax Constraints.

\end{itemize}

\subsection{Syntax Constraints}

\subsubsection{NYT Relation Set}

NYT is used by the NYT models in UniRel and DeepStruct.

There are 25 possible relations. The relations which most commonly occur in UniRel and DeepStruct output on FAA data are ``/location/location/contains", ``/location/neighborhood/neighborhood\_of", ``/business/person/company", and ``/business/company/place\_founded".

NYT relations are made up of three parts: the head entity is the subject, the middle part is a category for the head entity and the third part is a category or verb phrase that defines the relationship of the tail entity to the head. All triples where the head does not belong to the category in the middle part of the relation are docked 0.5 in syntactic accuracy. Similarly, if the last part of the relation is a category, such as ``place founded", then if the tail does not correspond to that category, i.e., is not a place, it is docked 0.5. Some relations have a verb phrase as the relation instead, such as ``contains". In this case, the tail must follow any implicit constraints of that verb. For example, the tail for ``contains" cannot be an abstract concept, since a location cannot contain an abstract concept. To summarize in an example: the relation /business/company/advisors must have a company as the head entity and a group of people as the tail.

\subsubsection{ACE-2005 Relation Set}

ACE-2005 is used by PL-Marker and DeepStruct. The ACE-2005 relations and their constraints on head and tail entities are laid out in Table \ref{tab:ACE-2005_Relations}.

\begin{table}[ht!]
\centering
\begin{adjustbox}{max width=\textwidth}
\begin{tabular}{|m{2cm}|m{4cm}|m{4cm}|}
  \hline
  \textbf{Relation} & \textbf{Head} & \textbf{Tail} \\ 
  \hline
  PER-SOC & Person(s) & Person(s) \\ 
  \hline
  ART & Person(s) & Physical Object \\ 
  \hline
  ORG-AFF & Person(s) & Organization \\ 
  \hline
  GEN-AFF & Any & Any \\ 
  \hline
  PHYS & Person(s) & Anything with a physical form \\ 
  \hline
  PART-WHOLE & Category must correspond to Tail & Category must correspond to Head \\ 
  \hline
\end{tabular}
\end{adjustbox}
\caption{ACE-2005 Relations' Syntax Constraints on  Head and Tail Entities}
\label{tab:ACE-2005_Relations}
\end{table}

\subsubsection{SciERC Relation Set}

SciERC is used by PL-Marker. The SciERC relations and their constraints are laid out in Table \ref{tab:SciERC_Relations}.

\begin{table}[ht!]
\centering
\begin{adjustbox}{max width=\textwidth}
\begin{tabular}{|m{3cm}|m{6cm}|m{6cm}|}
  \hline
  \textbf{Relation} & \textbf{Head} & \textbf{Tail} \\ 
  \hline
  PART-OF & Category must correspond to Tail & Category must correspond to Head \\ 
  \hline
  USED-FOR & Any & Any \\ 
  \hline
  FEATURE-OF & Category must correspond to Tail & Category must correspond to Head \\ 
  \hline
  CONJUNCTION & Category must correspond to Tail & Category must correspond to Head \\ 
  \hline
  HYPONYM-OF & Category must correspond to Tail & Category must correspond to Head \\ 
  \hline
  COMPARE & Category must correspond to Tail & Category must correspond to Head \\ 
  \hline
\end{tabular}
\end{adjustbox}
\caption{SciERC Relations' Syntax Constraints on Head and Tail Entities}
\label{tab:SciERC_Relations}
\end{table}

\subsubsection{REBEL Wikidata Property Relation Set}

REBEL utilizes a curated set of 220 relations derived from Wikidata Properties. The complete set is available in the REBEL repository.\footnote{The complete set of relations can be accessed in the REBEL repository at: \href{https://raw.githubusercontent.com/Babelscape/rebel/main/data/relations_count.tsv}{https://raw.githubusercontent.com/Babelscape/rebel/main/data/relations\_count.tsv}.}
The top 10 relations appearing in REBEL's output on the FAA data are: has part, part of, different from, subclass of, instance of, has effect, has cause, located in the administrative territorial entity, product or material produced, and facet of.

For all REBEL relations (Wikidata properties), the head and tail entities must match usage in Wikidata. This is judged by the evaluator. Some notable properties are:
\begin{itemize}
\item ``different from" is only used when head and tail entities share a similar name. The ``different from" relation is used to distinguish entities named the same way or similarly enough that they need to be distinguished.
\item ``has effect" and ``has cause" may have noun phrases or verb phrases on either end
\item ``has part", ``part of", ``subclass of", ``instance of", and ``facet of" all imply that the head and tail entities must correspond in entity category. We refer to coarsely defined categories such as event, physical object, time, quantity, date, and abstract concept, which are obvious to the annotator. If a categorical difference between head and tail is possible but not obvious, we count it as syntactically correct.
\end{itemize}

\subsection{Semantic Accuracy}
Semantic accuracy is the degree to which the tool's output adheres to the real world. A triple is either semantically accurate (1.0) or semantically inaccurate (0.0). The guidelines are recorded below:

\begin{itemize}
\item The evaluator is encouraged to use their domain expertise as well as all outside knowledge available.

\item If a head or tail entity is an incomplete phrase or includes extraneous words, the triple will still be counted as semantically accurate if using sub-spans of those entities enables a sensible triple. For example, in the record, ``Engine ran rough. Pilot landed in field," if the triple were (``engine", ``used by", ``pilot landed") were given, it would be counted as semantically accurate, since (``engine",``used by",``pilot") is accurate.
\end{itemize}

\subsection{Consistency}
Consistency is the degree to which the set of output triples for each record/document is free of contradictions. Percent consistency is calculated via the expression: ($N_{triples}$ - $N_{inconsistencies}$)/($N_{triples}$), where $N_{inconsistencies}$ is the number of triples such that if they were removed from the set of output triples, the remaining set would be consistent. For example, if there are 3 triples generated for a document and 2 of them contradict each other, there is 1 inconsistency since if one of the contradicting triples were removed, the remaining 2 would be consistent. In this case, it would receive a consistency score of 0.6667.

\begin{itemize}
    \item An example of contradicting triples would be (``Brookline, MA",``place of birth",``John F. Kennedy") and (``John F. Kennedy",``has place of birth",``Boston, MA")
    \item Most relations do not necessitate a one-to-one relation, however. In the record, ``Crashed when load wedged in trees. Improper preflight," if the triples (``improper preflight", ``has effect", ``crashed") and (``crashed",``has cause",``load wedged") were generated, this would still be consistent, since an event may have multiple causes.
\end{itemize}

\end{document}